\documentclass[12pt,onecolumn,draftcls,single spacing]{IEEEtran} %
\usepackage[utf8]{inputenc} 
\usepackage[T1]{fontenc}
\usepackage{url}
\usepackage{ifthen}
\usepackage{cite}
\usepackage[cmex10]{amsmath} %

\usepackage[
font=small,labelfont=bf
]{caption}
 \usepackage{flushend}
 \flushend
\usepackage{amsfonts, amssymb, amsthm, color,mathtools}
\usepackage{bbm}
\usepackage{comment}
\usepackage[inline]{enumitem}
\usepackage{thmtools}
\usepackage{thm-restate}

\usepackage{tikz}
\usepackage{diagbox}

\definecolor{DarkGreen}{rgb}{0.1,0.5,0.1}
\definecolor{DarkRed}{rgb}{0.5,0.1,0.1}
\definecolor{DarkBlue}{rgb}{0.1,0.1,0.5}
\usepackage[capitalize]{cleveref}
\usepackage{caption}
\usepackage{subcaption}
\usepackage{dsfont}
\usepackage[nolist]{acronym}
\usepackage{mathtools} %
\usepackage[scientific-notation=true]{siunitx}
\usepackage{float}
\usepackage{mathrsfs}

\IEEEoverridecommandlockouts

\usepackage{tcolorbox,tikz, lipsum}
\usetikzlibrary{backgrounds,calc,shadings,shapes.arrows,shapes.symbols,shadows,fit,positioning,topaths,hobby,shapes.geometric,shapes,snakes,patterns,arrows,automata,plotmarks}
\tcbuselibrary{skins}
\tcbuselibrary{breakable}
\usepackage{pgfplots}

\definecolor{VDarkGreen}{rgb}{0.05,0.25,0.05}
\makeatletter
\DeclareRobustCommand{\extstart}{%
  \@bsphack
  \leavevmode
  \color{black}%
  \@esphack
}
\DeclareRobustCommand{\extend}{%
  \@bsphack
  \normalcolor
  \@esphack
}
\makeatother

\usepackage[normalem]{ulem}
\usepackage{xspace}

\newcommand{\clusterindex}{\ensuremath{u}\xspace}
\newcommand{\nclients}{\ensuremath{n\xspace}}

\newcommand{\clusterclients}[1][\clusterindex]{\ensuremath{n_{#1}}\xspace}
\newcommand{\clientindex}[1][\clusterindex]{\ensuremath{i}\xspace}
\newcommand{\nclusters}{\ensuremath{b}\xspace}
\newcommand{\uecolluders}{\ensuremath{z_{\text{UE}}}\xspace}
\newcommand{\bscolluders}{\ensuremath{z_{\text{BS}}}\xspace}
\newcommand{\bsaddconn}{\ensuremath{v_{\clientindex}\xspace}}
\newcommand{\bsclient}[1][\clientindex]{\ensuremath{\mathcal{U}_{#1}}}
\newcommand{\bsclientone}{\ensuremath{\mathcal{U}_{\clientindex_1}}}
\newcommand{\bsclienttwo}{\ensuremath{\mathcal{U}_{\clientindex_2}}}

\newcommand{\I}[1]{\ensuremath{\mathrm{I}\left(#1\right)}}
\newcommand{\entropy}[1]{\ensuremath{\mathrm{H}\left(#1\right)}}

\newcommand{\observations}{\ensuremath{\mathcal{O}}\xspace}
\newcommand{\privatedata}{\ensuremath{\mathcal{G}}\xspace}
\newcommand{\dropouts}{\ensuremath{\mathcal{Z}}}
\newcommand{\aggregatedgradient}{\ensuremath{\mathbf{g}^\Sigma}\xspace}
\newcommand{\aggregatedkeys}{\ensuremath{\mathbf{k}^\Sigma}\xspace}

\newcommand{\model}[1][\clientindex]{\ensuremath{\mathbf{g}_{#1}}\xspace}
\newcommand{\submodel}[1][j]{\ensuremath{\mathbf{g}_{\clientindex}^{(#1)}}\xspace}

\newcommand{\randvec}[1][\clientindex]{\ensuremath{\mathbf{k}_{#1}}\xspace}
\newcommand{\subrandvec}[1][j]{\ensuremath{\mathbf{k}_{\clientindex}^{(#1)}}\xspace}
\newcommand{\graddim}{\ensuremath{d}\xspace}
\newcommand{\mincomm}{\ensuremath{C_\mathrm{LB}}\xspace}
\newcommand{\mincommr}{\ensuremath{C_\mathrm{LB,R}\xspace}}
\newcommand{\com}{\ensuremath{C}}
\newcommand{\uecom}{\ensuremath{C^{\text{UE}\rightarrow \text{BS}}}}
\newcommand{\bscom}{\ensuremath{C^{\text{BS}\rightarrow \text{BS}}}}
\newcommand{\bsfcom}{\ensuremath{C^{\text{BS}\rightarrow \text{F}}}}

\newcommand{\bsindex}{\ensuremath{\clusterindex}\xspace}
\newcommand{\tmpindex}{\ensuremath{j}\xspace}
\newcommand{\fold}{\ensuremath{a}\xspace}
\newcommand{\fnew}{\ensuremath{b}\xspace}
\newcommand{\foldevalstrunk}{\ensuremath{\mathbf{f}}\xspace}
\newcommand{\fullpaddedgrads}{\ensuremath{\mathbf{m}}}
\newcommand{\bspaddedgrads}[1][]{\ensuremath{{\fullpaddedgrads^{(#1)}}}}

\newcommand{\rpaddedgrads}[1][]{\ensuremath{{\tilde{\fullpaddedgrads}^{(#1)}}}}

\newcommand{\bssecrand}[1][]{\ensuremath{{\fullbssecrand^{(#1)}}}}

\newcommand{\fullbssecrand}{\ensuremath{\mathbf{r}}}
\newcommand{\rsecrand}[1][]{\ensuremath{{\tilde{\mathbf{r}}^{(#1)}}}}

\newcommand{\fullrsecrand}{\ensuremath{\mathbf{r}^\prime}}
\newcommand{\vanderbstrunk}{\ensuremath{\mathbf{V}}}
\newcommand{\eye}[1][]{\ensuremath{\mathbf{I}_{#1}}}
\newcommand{\vbsinvtrunk}{\ensuremath{\mathbf{N}}}
\newcommand{\vrb}[1][]{\ensuremath{\mathbf{D}^{(#1)}}}
\newcommand{\weightmat}[1][]{\ensuremath{{\mathbf{W}^{(#1)}}}}
\newcommand{\weightmatT}[1][]{\ensuremath{{\mathbf{W}^{(#1)T}}}}
\newcommand{\extrarand}[1][]{\ensuremath{\boldsymbol{\rho}_{#1}}}
\newcommand{\ftransport}[1][]{\ensuremath{t_{#1}}}
\newcommand{\partindex}{\ensuremath{\psi}}
\newcommand{\tsecrand}{\ensuremath{\boldsymbol{\gamma}}}

\newcommand{\veval}[1][\relayidx]{\ensuremath{\mathbf{v}_{{#1}}}}
\newcommand{\vevalr}[1][\relayidx]{\ensuremath{\mathbf{v}_{{#1}}^{\prime}}}

\newcommand{\tmprelayidxa}{\ensuremath{{m_1}}}
\newcommand{\tmprelayidxb}{\ensuremath{{m_2}}}
\newcommand{\contribution}[1][]{\ensuremath{\mathbf{c}_{#1}}}

\newcommand{\nbs}{\ensuremath{n}}
\newcommand{\kbs}{\ensuremath{k}}
\newcommand{\bssplits}{\ensuremath{\nu}}

\newcommand{\nr}{\ensuremath{{n^\prime}}}
\newcommand{\kr}{\ensuremath{{k^\prime}}}
\newcommand{\rsplits}{\ensuremath{{\nu^\prime}}}

\newcommand{\clientpoly}[1][x]{f_{\clientindex} (#1)}
\newcommand{\clientpolygrads}[1][x]{g_{\clientindex} (#1)}
\newcommand{\clientpolykeys}[1][x]{h_{\clientindex} (#1)}
\newcommand{\secrand}[1][]{\ensuremath{\mathbf{r}_{\clientindex}^{(#1)}}}
\newcommand{\secrandgrads}[1][]{\ensuremath{\mathbf{r}_{\clientindex}^{(#1)}}}
\newcommand{\secrandkeys}[1][]{\ensuremath{\mathbf{s}_{\clientindex}^{(#1)}}}

\newcommand{\sampleindex}{\ensuremath{l}}
\newcommand{\inputdim}{\ensuremath{d}}
\newcommand{\nuesamples}{\ensuremath{m_{\clientindex}}}
\newcommand{\nsamples}{\ensuremath{m}}
\newcommand{\uesample}{\ensuremath{\mathbf{x}_{\clientindex, \sampleindex}}}
\newcommand{\uelabel}{\ensuremath{y_{\clientindex, \sampleindex}}}
\newcommand{\uesamples}{\ensuremath{\mathbf{X}_{\clientindex}}}
\newcommand{\uelabels}{\ensuremath{\mathbf{y}_{\clientindex}}}
\newcommand{\samples}{\ensuremath{\mathbf{X}}}
\newcommand{\labels}{\ensuremath{\mathbf{y}}}
\newcommand{\lr}{\ensuremath{\eta}}
\newcommand{\iter}{\ensuremath{t}}
\newcommand{\globalmodel}{\ensuremath{\mathbf{w}}}

\newcommand{\figheight}{2.3in}
\newcommand{\figwidth}{4in}

\newcommand{\clientset}{\ensuremath{\mathcal{N}}}
\newcommand{\noncuriousset}{\ensuremath{\mathcal{H}}}
\newcommand{\uecolluderset}{\ensuremath{\mathcal{C}}}
\newcommand{\bscolluderset}{\ensuremath{\mathcal{B}}}

\newcommand{\Submodels}[1][\mathcal{N}]{\ensuremath{\mathcal{G}^{#1}}\xspace}
\newcommand{\Obs}{\ensuremath{\mathcal{F}^{\bscolluderset}}\xspace}
\newcommand{\Randvecs}[1][\clientset]{\ensuremath{\mathcal{K}^{#1}}\xspace}
\newcommand{\Randvecscol}{\ensuremath{\mathcal{K^\mathcal{\uecolluderset}}}\xspace}
\newcommand{\Sumvecs}{\ensuremath{\mathcal{S}}\xspace}
\newcommand{\defeq}{\mathrel{\mathop:}=}
\newcommand{\lossf}{\ensuremath{L}}
\newcommand{\fed}{F}

\newcommand{\ext}[1]{{\color{black} #1}}

\newcommand{\ntotalrelays}{\ensuremath{r}}
\newcommand{\relaycollusions}{\ensuremath{z_\text{R}}}
\newcommand{\relayidx}{\ensuremath{m}}

\newcommand{\bsset}{\ensuremath{\mathcal{E}_{\clientindex}}}
\newcommand{\relayset}{\ensuremath{\mathcal{M}_{\clientindex}}}

\newcommand{\bseval}{\ensuremath{\alpha}}
\newcommand{\relayeval}{\ensuremath{{\beta}}}

\newcommand{\bsrcom}{\ensuremath{C^{\text{BS}\rightarrow \text{R}}}}
\newcommand{\rcom}{\ensuremath{C^{\text{R}\rightarrow \text{R}}}}
\newcommand{\rfcom}{\ensuremath{C^{\text{R}\rightarrow \text{F}}}}
\newcommand{\relaybs}{\ensuremath{\mathcal{J}_\clusterindex}}

\newcommand{\tmpconn}{\ensuremath{\mathcal{U}}}
\newcommand{\tmpmain}{\ensuremath{\mathcal{Y}}}
\newcommand{\tmpsec}{\ensuremath{\mathcal{X}}}
\newcommand{\tmpbs}{\ensuremath{\mathcal{E}}}
\newcommand{\tmprelay}{\ensuremath{\mathcal{M}}}
\newcommand{\mainsets}{\ensuremath{\mathscr{Y}}}
\newcommand{\secsets}{\ensuremath{\mathscr{X}}}
\newcommand{\bsmain}{\ensuremath{\mathcal{Y}_{\clientindex}}}
\newcommand{\bssec}{\ensuremath{\mathcal{X}_{\clientindex}}}
\newcommand{\common}{\ensuremath{\mathcal{P}}}
\newcommand{\commonmain}{\ensuremath{\common^\tmpmain}}
\newcommand{\commonsec}{\ensuremath{\common^\tmpsec}}

\newcommand{\commonconn}[1][\mathcal{U}]{\ensuremath{\common^{#1}}}
\newcommand{\commonbs}[1][\mathcal{B}]{\ensuremath{\common^{#1}}}
\newcommand{\commonrelay}[1][\mathcal{M}]{\ensuremath{\common^{#1}}}

\newcommand{\lbsset}{\ensuremath{b_{\clientindex}}}
\newcommand{\lrelayset}{\ensuremath{r_{\clientindex}}}

\newcommand{\maxcard}{\ensuremath{w}}
\newcommand{\maxcolluders}{\ensuremath{z}}
\newcommand{\cardset}{\ensuremath{k_{\clientindex}}}

\newcommand{\tmpset}{\ensuremath{\mathcal{T}}}
\newcommand{\bsaddconmain}{\ensuremath{y_{\clientindex}}}
\newcommand{\bsaddconsec}{\ensuremath{x_{\clientindex}}}

\newcommand{\gradpolyset}{\ensuremath{\mathcal{A}}}
\newcommand{\keypolyset}{\ensuremath{\mathcal{Q}}}
\newcommand{\setgradientsums}{\ensuremath{\mathcal{W}}}
\newcommand{\setpaddedsums}{\ensuremath{\mathcal{V}}}

\newcommand{\tsssecret}{\ensuremath{M}}
\newcommand{\tssdim}{\ensuremath{{d_M}}}
\newcommand{\tssshare}{\ensuremath{S}}

\newcommand{\tsspmf}{\ensuremath{p_\tsssecret}}
\newcommand{\tsssharedim}{\ensuremath{\ell}}
\newcommand{\given}{\mid}
\newcommand{\tssind}{\ensuremath{i}}
\newcommand{\tssinds}{\ensuremath{\mathcal{I}}}
\newcommand{\tsstmpset}{\ensuremath{\mathcal{X}}}
\newcommand{\tssn}{{n_\mathrm{s}}}
\newcommand{\tssk}{{k_\mathrm{s}}}
\newcommand{\tssz}{{z_\mathrm{s}}}

\newtheorem{theorem}{Theorem}
\newtheorem{proposition}{Proposition}

\newtheorem{example}{Example}
\newtheorem{remark}{Remark}

\theoremstyle{definition}
\definecolor{britishracinggreen}{rgb}{0.0, 0.26, 0.15}
\newtheorem{definition}{Definition}

\crefname{paragraph}{paragraph}{paragraphs}
\Crefname{paragraph}{Paragraph}{Paragraphs}

\title{Private Aggregation in Hierarchical Wireless Federated Learning with Partial and Full Collusion}

\author{%
  \IEEEauthorblockN{
                    Maximilian Egger,
                    Christoph Hofmeister,
                    Antonia Wachter-Zeh
                    and Rawad Bitar
                    }
                    
  \IEEEauthorblockA{%
                     \normalsize	Institute for Communications Engineering, Technical University of Munich,
                    Germany \\
                    \{maximilian.egger, christoph.hofmeister, antonia.wachter-zeh, rawad.bitar\}@tum.de \vspace{-1.5cm}
                    }
                     
\thanks{This project has received funding from the German Research Foundation (DFG) under Grant Agreement Nos. BI 2492/1-1 and WA 3907/7-1 and from the European Research Council (ERC) under the European Union’s Horizon 2020 research and innovation programme (grant agreement No. 801434).
\ext{Preliminary results were presented at IEEE International Symposium on Information Theory (ISIT) 2023~\cite{egger2023private}.}}
}

\begin{document}

\maketitle
\begin{abstract}
In federated learning, a federator coordinates the training of a model, e.g., a neural network, on privately owned data held by several participating clients. The gradient descent algorithm, a well-known and popular iterative optimization procedure, is run to train the model. Every client computes partial gradients based on their local data and sends them to the federator, which aggregates the results and updates the model. Privacy of the clients' data is a major concern. In fact, it is shown that observing the partial gradients can be enough to reveal the clients' data. 
\extstart 
Existing literature focuses on private aggregation schemes that tackle the privacy problem in federated learning in settings where all users are connected to each other and to the federator. 
\extend
In this paper, we consider a hierarchical wireless system architecture in which the clients are connected to base stations; the base stations are connected to the federator either directly or through relays. We examine 
\extstart settings with and without relays,  \extend
and derive fundamental limits on the communication cost under information-theoretic privacy with different collusion assumptions. We introduce suitable private aggregation schemes tailored for these settings whose communication costs are multiplicative factors away from the derived bounds.
\end{abstract}


\section{Introduction}
Federated learning consists of a set of clients collaboratively training a neural network on private local data through the help of a central entity, referred to as the federator \cite{mcmahan2017communication}. The training process is based on optimization procedures such as gradient descent, which iteratively updates the parameters of a model (e.g., a neural network) subject to gradient computations with respect to a differentiable loss function. To avoid moving the clients' data to the federator, at every iteration of the gradient descent algorithm each client evaluates the gradient of the loss function on their private data and sends the result to the federator. The federator then aggregates the partial computations to update the global model parameters.

In the original proposal of federated learning \cite{mcmahan2017communication}, each of the clients directly sends their computational results (i.e., the local model updates) to the federator. However, this gives rise to model inversion attacks that can possibly allow the federator to learn the clients' private data \cite{jeon2021gradient}. Since privacy of the clients' data is of paramount importance, the concept of secure aggregation (to which we shall refer as \emph{private aggregation}) has been introduced in \cite{bonawitz2017practical}. Ideas from secret sharing \cite{shamir1979how,blakley1979safeguarding,mceliece1981sharing} and secure multi-party computation \cite{cramer2015secure} are used to aggregate the partial computations of the clients in a privacy-preserving manner. The federator is guaranteed to only observe the aggregation of all partial computations, i.e., only the information necessary to update the global model parameters of the neural network. 

A simple version of private aggregation works as follows. Each client generates a random vector and adds it to their partial computation. Only one client subtracts the sum of all random vectors used by the other clients from their partial computation. Observing individual vectors, the federator learns no information about the clients' computation. However, the federator obtains the desired sum of all partial computations by aggregating all received vectors, since the random vectors cancel out. This scheme, and all schemes based on secret sharing, suffer from the presence of slow or unresponsive clients termed \emph{stragglers}. For example, if a single client does not send their partial computation to the federator, the whole iteration has to be discarded.

Constructing private aggregation schemes that can mitigate the effect of stragglers and tolerate clients dropping out from the learning process, called \emph {dropouts}, is an active area of research, cf.  \cite{jahani2022swiftagg+,so2022lightsecagg,zhao2021information,schlegel2021codedpaddedfl,li2020federated,kairouz2021advances}. \extstart However, \extend the proposed schemes consider settings in which communication links between all the clients and direct communication links between the clients and the federator exist. 
In this work, we consider a hierarchical wireless network architecture in which the clients are connected to each other only through base stations. We consider two cases: \emph{(a)} the base stations are directly connected to the federator; and \emph{(b)} the base stations can transmit information to the federator only through intermediate relay nodes. Such settings are considered in, e.g., \cite{ye2023heterogeneous,abad2020hierarchical,chen2020semi,demir2021foundations, chen2022survey}. Directly applying an existing scheme \extstart does not fulfill the required privacy guarantees, since base stations, wireless service providers, and potential intermediate nodes, obtain enough information to decode the clients' sensitive computations. To apply existing schemes, one must rely on computationally private encryption methods instead. \extend We design novel \extstart information-theoretically private \extend aggregation schemes tailored to such wireless system architectures. The scheme guarantees the privacy of the clients' data against base stations, relay nodes, the federator, and other clients.

The proposed methods require the clients to be connected to multiple base stations. This can, for example, be ensured by leveraging client movements, intersections in transmission ranges of base stations, and the recently proposed wireless architecture of user-centric massive MIMO \cite{demir2021foundations, chen2022survey} that goes beyond conventional system architectures. 
In these systems, clients are surrounded and likely outnumbered by base stations.

In this work, we consider information-theoretic privacy guarantees. Ensuring this strong privacy guarantee requires embedding the local computations in finite fields, which necessitates some quantization and is an inherent restriction for system design. While quantization introduces additional noise and affects the learning process, it is used in many gradient compression schemes to significantly reduce the communication cost. \extstart It has been shown \extend that gradient descent with gradient compression schemes such as 1-bit Stochastic Gradient Descent (SGD) \cite{seide2014onebit}, Quantized SGD \cite{karimireddy2019error}, TernGrad \cite{wen2017terngrad}, sign-based SGD with error feedback \cite{karimireddy2019error}, Vector Quantized SGD \cite{gandikota2021vqsgd} and Natural Compression \cite{horvoth2022natural} retains satisfactory final model accuracy even with aggressive quantization.

\emph{Related work:} %
Private aggregation for federated learning is receiving significant attention from the scientific community. The authors of the primal work~\cite{bonawitz2017practical} first introduced a protocol with quadratic communication cost $\mathcal{O}(\nclients^2)$ in the number of participating clients $\nclients$ that is capable of tolerating stragglers. Follow-up works on private aggregation improve the communication cost while tolerating a certain fraction of stragglers. For example, in \cite{bell2020secure}, the communication cost is reduced to a polylogarithmic overhead, whereas the protocol in~\cite{so2021turbo} called Turbo-Aggregate reduces the communication to $\mathcal{O}(\nclients \log \nclients)$.
In contrast, the scheme termed FastSecAgg \cite{kadhe2020fastsecagg} uses a novel secret sharing scheme to reduce the computation cost at the server while, under some conditions on the relation of the number of model parameters over the number of clients, maintaining similar communication complexity compared to prior schemes. LightSecAgg~\cite{so2022lightsecagg} tackles the problem of increasing complexities with increasing dropout resilience. Closest to our work is the scheme termed SwiftAgg \cite{jahani2022swiftagg}, which reduces the communication load at the federator. SwiftAgg+ \cite{jahani2022swiftagg+} is a generalization of SwiftAgg that flexibly reduces the communication load depending on the network conditions, still requiring client-to-client connections. For example, for fully connected networks of clients, the communication cost reaches its minimum of $\nclients (1+\frac{z+d}{\nclients-z-d})$, where $z$ is the number of colluding clients and $d$ is the maximum tolerable number of users that drop out.
Other recent works on private aggregation under different system models include, e.g., \cite{schlegel2023coded,sami2023secure}, and further demonstrate the interest in such methods. We refer interested readers to the surveys \cite{li2020federated,kairouz2021advances} for more details and a comprehensive overview.

\emph{Contributions:} Our main contribution is designing and analyzing an information-theoretically private aggregation scheme for hierarchical wireless federated learning, where clients are connected to the federator on the wireless links only through base stations. We consider partial collusion in which the federator does not collude with the base station and full collusion in which all entities can collude. We derive fundamental limits on the communication cost of information-theoretic private aggregation schemes and compare them to that of our proposed scheme. We analyze the impact of the additional level of intermediate nodes in the hierarchy, through which the base stations are connected to the federator. We derive fundamental limits for both collusion assumptions and provide a suitable private aggregation scheme. We utilize tools from secret sharing \cite{shamir1979how,mceliece1981sharing} and show that the communication costs of our designed schemes are a small multiplicative factor away from the derived bounds.

\emph{Outline:} We introduce preliminaries and notation in \cref{sec:prelims}. In \cref{sec:system_architecture}, we describe the wireless system architecture for which we derive fundamental limits on communication cost under privacy constraints. We propose in \cref{sec:it_priv} and \cref{subsec:arbitrary_colluders} suitable private aggregation schemes for both collusion assumptions, respectively, and analyze their communication complexity. In \cref{sec:extended_scheme}, we describe a more sophisticated system model with an additional layer of hierarchy that requires the base station to transmit to the federator via relaying nodes. We derive a fundamental lower bound on the communication cost and present a scheme that, under mild assumptions, matches this bound up to a small multiplicative factor. Lastly, in \cref{sec:transform}, we discuss how to transform shares of a secret sharing scheme into shares of another scheme with different parameters, which is required if the properties or requirements change across different levels of the system hierarchy. We conclude in \cref{sec:conclusion}.

\section{Preliminaries} \label{sec:prelims}

\emph{Notation:} Sets are denoted by calligraphic letters. Vectors are denoted by bold lower case letters, and matrices by bold upper case letters. Unless stated otherwise, vectors are row vectors. The set of positive integers $\{1,\dots,n\}$ is denoted by $[n]$. We use the same letters for random variables and their realizations, which can be distinguished by the context. $\mathrm{I}(X;Y)$ denotes the mutual information between random variables $X$ and $Y$. $\mathrm{H}(X)$ denotes the entropy of $X$.

\emph{Distributed Stochastic Gradient Descent:}
We consider the problem of optimizing a predictive model\footnote{This work applies to any problem setting in which the federator wants to aggregate the contributions of the clients, such as federated averaging.}, e.g., a neural network, based on the well-known stochastic gradient descent {algorithm}. Every client $\clientindex \extstart \in [n]\extend$ owns a data set, containing $\nuesamples$ samples $\uesample \in \mathbb{R}^{1\times \inputdim}, \sampleindex \in [\nuesamples],$ and their corresponding labels $\uelabel \in\mathbb{R}$. By stacking the samples $\uesample$ {row-wise}, we obtain the data matrix $\uesamples \in \mathbb{R}^{\nuesamples \times \inputdim
}$ of client $\clientindex$. The vector containing all $\nuesamples$ labels reads as $\uelabels\in\mathbb{R}^{\nuesamples}$. The entire data set consequently consists of $\nsamples = \sum_{\clientindex=1}^\nclients \nuesamples$ samples and is given by the matrix of features $\samples \in \mathbb{R}^{\nsamples \times \inputdim}$ and the vector of labels $\labels \in \mathbb{R}^{\nsamples}$, obtained through row-wise stacking of the clients' sample matrices $\uesamples$ for all clients $\clientindex \in [\nclients]$ and the vectors $\uelabels$, respectively. %

Given a function $f(\mathbf{x}, \globalmodel)$, the goal is to compute values for the model parameters $\globalmodel$, such that
$f(\mathbf{x}, \globalmodel)$ appropriately predicts the label $y$ of a sample $\mathbf{x}$, i.e., $f(\mathbf{x}, \globalmodel) \approx y$.
The predictions are called satisfactory if a given differentiable loss function $\lossf(\samples, \labels, \globalmodel)$ has a small value, i.e., the goal is to find an exact or approximate solution to the optimization problem $\arg\min_{\globalmodel} \lossf(\samples, \labels, \globalmodel)$.
In gradient descent, this is done by iteratively updating the model parameters $\globalmodel$. Starting with initial model parameters $\globalmodel_0$, at every iteration $\iter>0$ the model is updated according to
\vspace{-.15cm}
\begin{align*}
    \globalmodel_\iter &= \globalmodel_{\iter-1} - \frac{\lr}{\nsamples} \nabla_{\globalmodel} \lossf(\samples, \labels, \globalmodel_{\iter-1}) \\
    &= \globalmodel_{\iter-1} - \frac{\lr}{\nsamples} \sum_{\clientindex=1}^\nclients \nabla_{\globalmodel} \lossf(\uesamples, \uelabels, \globalmodel_{\iter-1}),
\end{align*}

\vspace{-.15cm}
\noindent where for the latter equation we assume that the loss function is additively separable, i.e., $\lossf(\samples, \labels, \globalmodel) = \sum_{\clientindex=1}^\nclients \sum_{\sampleindex=1}^{\nuesamples} \lossf(\uesample, \uelabel, \globalmodel)$. Once clients drop out at a certain iteration, we obtain a stochastic estimate of the true gradient, referred to as stochastic gradient descent. Let $\dropouts$ be the set of indices corresponding to clients who did not contribute their partial gradient $\nabla_\globalmodel \lossf(\uesamples, \uelabels, \globalmodel_\iter)$ at iteration $\iter$, then the update function is given by
\vspace{-.1cm}
\begin{align*}
    \globalmodel_\iter &= \globalmodel_{\iter-1} - \frac{\lr}{\sum_{\clientindex \in [\nclients]\setminus \dropouts} \nuesamples} \sum_{\clientindex\in [\nclients]\setminus\dropouts} \nabla_{\globalmodel} \lossf(\uesamples, \uelabels, \globalmodel_{\iter-1}).
\end{align*}

\noindent If the individual \emph{partial gradients} $\mathbf{g}_{\clientindex,\iter} \defeq \nabla_{\mathbf{w}} \lossf(\uesamples, \uelabels, \mathbf{w}_t)$ computed by the clients who did not drop out at iteration $\iter$ are sent to the federator, the stochastic update can be computed through a weighted linear combination of the $\mathbf{g}_{\clientindex,\iter}$'s, for all $\clientindex\in [\nclients]\setminus\dropouts$. For the optimization process, however, it is sufficient if the federator knows the sum $\sum_{\clientindex\in [\nclients]\setminus\dropouts} \mathbf{g}_{\clientindex,\iter}$. Ensuring that the federator only learns the aggregate leads to a better level of privacy for the clients' data. The private aggregation happens at every iteration. For clarity of presentation, we consider one particular iteration and drop the iteration index $\iter$, i.e., we refer to the gradient of client $\clientindex$ as $\model$. In this work, we do not focus on dropout mitigation, but our scheme is compatible with standard approaches as, e.g., in \cite{yu2019lagrange,jahani2022swiftagg+}.

\extstart \paragraph*{Secret Sharing}
In order to achieve information-theoretic privacy, we make use of secret sharing, which we define below.
\begin{definition}[$(\tssn, \tssk, \tssz)$ Threshold Secret Sharing]
  Let $0 < \tssz < \tssk \leq \tssn$ be integers and let $\tsssecret \in \mathbb{F}_q^\tssdim$ be a discrete random variable with probability mass function $\tsspmf$ representing a private message. 
  A randomized encoding function that outputs a collection of $\tssn$ discrete random variables called shares, denoted by $\tssshare_1, \dots, \tssshare_\tssn$, is called an $(\tssn, \tssk, \tssz)$ threshold secret sharing of $\tsssecret$ if for any admissible $\tsspmf$ it holds that: \begin{enumerate*}[label={\textit{\arabic*)}}]
    \item any $\tssk$ shares $\tssshare_{i_1},\ldots, \tssshare_{i_k}$ fully determine the value of $\tsssecret$, i.e., $\entropy{M \mid \tssshare_{i_1},\dots, \tssshare_{i_\tssk}} = 0$; and
    \item any $\tssz$ shares $\tssshare_{i_1},\ldots, \tssshare_{i_z}$ reveal no information about the value $\tsssecret$, i.e., $\entropy{\tsssecret \mid \tssshare_{i_1},\dots, \tssshare_{i_\tssz}} = \entropy{\tsssecret}$.
\end{enumerate*}
\end{definition}

Our bounds on the communication cost of any wireless private aggregation scheme depend on the minimum share size of a secret sharing. Although not novel, e.g.~\cite{huang2016communication}, we state the result in our notation here and prove it in~\cref{app:proof_of_sharesize} for completeness.
\begin{proposition}[Minimum Combined Share Size] \label{prop:sharesize}
  For any $(\tssn, \tssk, \tssz)$ secret sharing of a secret $\tsssecret \in \mathbb{F}_q^\tssdim$ with $\tssshare_1 \in \mathbb{F}_q^{\tsssharedim_1}, \dots, \tssshare_\tssn \in \mathbb{F}_q^{\tsssharedim_\tssn}$ it holds that \begin{align*}
    \sum_{i\in [\tssn]} \tsssharedim_i \geq \tssdim \frac{\tssn}{\tssk-\tssz}.
  \end{align*}
\end{proposition}
\extend

\section{Wireless System Architecture \extstart Without Relays \extend} \label{sec:system_architecture}

We \ext{first} consider the setting in which a \emph{federator} is connected to $\nclusters$ \emph{base stations}. The \emph{user equipment} (UEs) of the \extstart $\nclients$ \extend participating \emph{clients} are connected \extstart via point-to-point links to the base stations, which in turn are connected to the federator. Each client $\clientindex$'s UE is connected to a certain set of base stations $\bsclient \subseteq [\nclusters]$ of cardinality $\vert \bsclient \vert = \clusterclients[\clientindex]$. \extend The system architecture is depicted in~\cref{fig:systemmodel}.

\begin{figure}[t]
\centering
    \resizebox{0.7\linewidth}{!}{\input{resources/system_model_no_clusters}}
    \caption{
    The wireless architecture for federated learning in the simplified setting. \extstart Arrows correspond to point-to-point links. \extend The user equipments of the clients (termed UEs) are connected to the federator and to each other via base stations. Each UE is connected to several base stations. Each client is connected to at least $\bscolluders+1$ base stations. %
    }
    \label{fig:systemmodel} \vspace{-0.5cm}
\end{figure}

\emph{Adversary Model:} For all entities, i.e., the federator, the base stations, and the clients, we consider an honest-but-curious adversary model. The strongest collusion assumption allows at most $\uecolluders$ clients to collude with at most $\bscolluders$ base stations and with the federator. Colluding here refers to the entities sharing their observed data to infer information about other clients' private data. \extstart Note that also parties not directly connected to each other can collude. \extend Several measures must be taken to ensure such a privacy requirement. For clarity of exposition, we first consider a partial collusion assumption that is of independent interest. We construct a private aggregation scheme for this assumption and then extend it to the full collusion assumption. The partial collusion does not allow the federator to collude with the base stations, i.e., at most $\uecolluders$ clients are allowed to collude (a) with at most $\bscolluders$ base stations \emph{or} (b) with the federator. The modification of our scheme to cover this scenario will later be discussed in \cref{remark:collusions} and investigated in \cref{subsec:arbitrary_colluders}.

\emph{Privacy Measure:} 
We construct and analyze a scheme that ensures information-theoretic privacy.
To cope with this privacy requirement, we assume that all gradients live in a finite field, i.e., $\model \in \mathbb{F}_q^{\inputdim}$ for some prime power $q$, which can be ensured by proper quantization and mapping to a sufficiently large finite field as is common in the literature, e.g., \cite{yu2019lagrange,jahani2022swiftagg+}. Let $\mathcal{C}$ and $\mathcal{B}$ denote the sets of all $\uecolluders$ colluding clients and all $\bscolluders$ colluding base stations, respectively. We abuse notation and let $\observations^\mathcal{C}$, $\observations^\mathcal{B}$ and $\observations^\fed$ denote the collection of random variables representing the data observed by the clients in $\mathcal{C}$, base stations in $\mathcal{B}$ and the federator, respectively. Let $\tmpset$ be a set of clients; we denote by $\Submodels[\tmpset]$ the collection of the gradient vectors of the clients in $\tmpset$. Furthermore, let $\clientset \defeq [\nclients]$ be the set of all clients, and $\noncuriousset \defeq \clientset \setminus \uecolluderset$ be the set of honest and non-curious clients. Then, we can denote by $\aggregatedgradient$ the value of the aggregated gradients of all clients in the set $\noncuriousset$ of honest clients.
With this notation\extstart~and following the literature on private aggregation, e.g.~\cite{jahani2022swiftagg+, so2022lightsecagg, sami2023secure}\extend, we formally define the privacy guarantee \extstart when all types of entities are allowed to collude (\emph{full collusion}) as \extend
\begin{align*}
    &\mathrm{I}\left(\observations^{\mathcal{C}}, \observations^\mathcal{B}, \observations^F; \privatedata^{\noncuriousset} \given \aggregatedgradient, \privatedata^{\uecolluderset} \right) = 0.
\end{align*}

The privacy guarantee under the partial collusion assumption described by cases (a) and (b) can be formalized as in \eqref{eq:it_privacy} and \eqref{eq:it_privacy_federator}, respectively. In the sequel, we consider schemes that satisfy both \extstart types of \extend assumptions.
\begin{align}
    &\mathrm{I}\left(\observations^{\mathcal{C}}, \observations^{\mathcal{B}}; \privatedata^{\noncuriousset} \given \extstart\aggregatedgradient, \extend \privatedata^{\uecolluderset} \right) = 0 \text{ and } \label{eq:it_privacy}\\
    &\mathrm{I}\left(\observations^{\mathcal{C}}, \observations^F; \privatedata^{\noncuriousset} \given \aggregatedgradient, \privatedata^{\uecolluderset} \right) = 0. \label{eq:it_privacy_federator} %
\end{align}
 
Note that %
the privacy guarantee is conditioned on the knowledge of $\aggregatedgradient$ since it is required by the federator for the operation of the learning algorithm and will be known to the clients at the next iteration. %
Since \extstart any information leakage of $\privatedata^{\uecolluderset}$ about $\privatedata^{\noncuriousset}$ is fundamentally unavoidable, \extend conditioning on $\privatedata^{\uecolluderset}$ represents the strongest information-theoretic privacy for this situation. 
\extstart Intuitively, the privacy guarantee means that the colluders cannot learn any additional information about the honest clients' data beyond what is necessary for the scheme to operate and beyond what the colluding clients already know about the honest clients' data before the scheme. \extend %
Given the relaxed privacy requirements, we derive in \cref{thm:converse} a fundamental lower bound on the total cost of communication $\mincomm$ incurred by any scheme guaranteeing information-theoretic privacy as in cases (a) and (b). The communication cost is defined as follows. 

\emph{Communication Cost:} The communication cost of a scheme is the number of symbols in $\mathbb{F}_q$ required per iteration to privately aggregate and transmit the aggregated gradient from the clients to the federator through the base stations. The cost of distributing the global model is a constant irrespective of the scheme and hence omitted for clarity of presentation.

\begin{theorem}[Lower Bound on the Communication Cost] \label{thm:converse}
Assume a wireless federated learning architecture as depicted in~\cref{fig:systemmodel} where each client $\clientindex \in [\nclients]$ is connected to \mbox{$\bscolluders + \bsaddconn \leq \nclusters$} base stations, $v_i>0$.
Any scheme that guarantees information-theoretic privacy, for all gradient distributions, against $\uecolluders$ clients colluding with $\bscolluders$ base stations (cf.~\eqref{eq:it_privacy}) or with the federator (cf.~\eqref{eq:it_privacy_federator}) requires a communication cost $\com$ bounded from below by
\begin{align*}
    \com \geq \mincomm \defeq \graddim \left(\max_{\clientindex \in [\nclients]}\left\{ \frac{\bscolluders+\bsaddconn}{\bsaddconn}\right\} + \sum_{\clientindex=1}^\nclients \frac{\bscolluders + \bsaddconn}{\bsaddconn} \right). %
\end{align*}
\end{theorem}
\begin{proof}
We give lower bounds on the communication cost of private aggregation schemes guaranteeing privacy as in~\eqref{eq:it_privacy} and~\eqref{eq:it_privacy_federator} separately. A scheme \extstart fulfilling \extend both guarantees must have a communication cost of at least the maximum of the derived bounds. Both bounds assume the gradients of the clients to be independent and uniformly distributed over $\mathbb{F}_q^\inputdim$, i.e., the gradients cannot be compressed any further.

For schemes with privacy against $\uecolluders$ \extstart clients \extend colluding with the federator \eqref{eq:it_privacy_federator}, the bound on the communication cost degenerates to the non-private case, i.e., $\com \geq \graddim (\nclients + \nclusters^\prime)$, where $\nclusters^\prime \leq \nclusters$ is the cardinality of the smallest set of base stations such that each client $\clientindex \in [\nclients]$ is connected to at least one base station.

For schemes guaranteeing privacy against $\uecolluders$ clients colluding with $\bscolluders$ base stations \eqref{eq:it_privacy}, the bound proceeds by a counting argument. Every client $\clientindex$ computes a gradient of dimension $\graddim$, which has to be communicated to the federator. Due to the distribution of the gradients, no lossless compression is possible. Hence, each client must communicate at least a vector of dimension $\graddim$. While communication with other clients can be avoided to overcome the problem of collusions, the only way for clients to communicate with the federator is through base stations. 
To successfully communicate the clients' gradients while guaranteeing privacy against $\bscolluders$ colluding base stations, the messages sent from the clients to the base stations must form \extstart an $(\tssn=\bscolluders+\bsaddconn, \tssk=\bscolluders+\bsaddconn, \tssz=\bscolluders)$ secret sharing of $\model[\clientindex]$. \extend The total client to base station communication is minimized when each client sends shares to all $\bscolluders + \bsaddconn$ connected base stations and the minimum \extstart total \extend communication from each client $\clientindex$ to \extstart its connected base stations is $\graddim \frac{\bscolluders+\bsaddconn}{\bsaddconn}$ by~\cref{prop:sharesize}. \extend
We sum over all clients to obtain the right part of the lower bound shown in the statement of the theorem. The minimum communication required to transmit the clients' results from the base stations to the federator is determined by the client with the worst connectivity. Consequently, the base stations collectively have to transmit at least a vector of dimension $\max_{\clientindex \in[\nclients]} \extstart \graddim \extend \frac{\bscolluders + \bsaddconn}{\bsaddconn}$. 
\end{proof}

\section{Private Aggregation with Partial Collusion} \label{sec:it_priv}
We construct a private aggregation scheme for wireless federated learning using the architecture depicted in~\cref{fig:systemmodel} that acts in two phases: a) secret key generation and secret sharing through the base stations; and b) secret key aggregation. %

We assume that each client $\clientindex \in [\nclients]$ is connected to at least $\bscolluders+1$ base stations. Otherwise, information-theoretic privacy against the base stations cannot be ensured for that client.\footnote{If certain clients are not connected to more than $\bscolluders$ base stations or information-theoretic privacy is not required, our scheme can be designed for computational privacy. In phase a), the federator distributes secret keys to those clients using public key cryptography. In phase b), the client's results are padded using the output of a cryptographic pseudo-random number generator where the secret keys are used as seed, i.e., a stream cipher. The communication cost for those clients approaches the cost without privacy.}
 
\paragraph{Secret Key Generation and Secret Sharing Through the Base Stations} \label{para:ss_through_bs}
At the start of each iteration, each client $\clientindex$ \extstart independently \extend generates a vector $\randvec[\clientindex]$ (called key) uniformly at random from $\mathbb{F}_q^\graddim$, i.e., $\randvec[\clientindex]$ is of the same dimension as the gradient vector $\model$. On a high level, every client pads (mixes) its private gradient with the random vector and uses secret sharing to distribute it to $\clusterclients[\clientindex] > \bscolluders$ base stations. To that end, the gradient $\model$ and the random vector $\randvec$ are split into $\bsaddconn \defeq \clusterclients[\clientindex] - \bscolluders$ parts\footnote{We assume that $\bsaddconn$ divides $\graddim$, which can be ensured by zero-padding.}, i.e., 
$\model = (\submodel[1] \dots\ \submodel[\bsaddconn])$, and $\randvec = (\randvec^{(1)} \dots\ \randvec^{(\bsaddconn)})$.
The secret shares are generated by encoding the vector $\model + \randvec[\clientindex]$ into a polynomial 
\vspace{-.1cm}
\begin{equation}
    \clientpoly = \sum_{j \in [\bsaddconn]} x^{j-1} (\submodel[j] + \subrandvec[j]) + \sum_{j \in [\bscolluders]} x^{\bsaddconn+j-1} \secrand[j], \label{eq:clientpoly}
\end{equation}

\vspace{-.2cm}
\noindent where $\forall j \in [\bscolluders], \secrand[j] \in \mathbb{F}_q^{\graddim/\bsaddconn}$ are random vectors whose entries are drawn independently and uniformly at random from $\mathbb{F}_q$.

Each base station $\clusterindex$ is assigned a distinct evaluation point $\alpha_{\clusterindex}$. Every client $i$ then sends the share $\clientpoly[\alpha_{\clusterindex}]$ to base station $\clusterindex$ for all $\clusterindex \in \bsclient$. To be able to decode, i.e., to obtain the vector $\model + \randvec$, the federator should receive $\bscolluders+\bsaddconn$ evaluations of the polynomial $\clientpoly$. The base stations can simply forward the message $\clientpoly[\alpha_{\clusterindex}]$ to the federator who can then interpolate the received polynomials and thus obtain $\sum_{\clientindex \in [\nclients]} (\model[\clientindex] + \randvec[\clientindex]$).

The communication cost of the scheme is further reduced as follows. If a base station receives evaluations from a set of clients $\commonconn \subseteq [\nclients]$ that share the same connectivity pattern $\tmpconn$, i.e., for all $i,j \in \commonconn$, $i\neq j$, it holds that $\bsclient[i] = \bsclient[j] = \tmpconn$, the base station $\clusterindex$ can sum the evaluations corresponding to clients $\clientindex \in \commonconn$ and send the aggregation $\sum_{\clientindex \in \commonconn} \clientpoly[\alpha_{\clusterindex}]$ to the federator. %

For every connectivity pattern $\tmpconn \in \{\bsclient\}_{\clientindex \in [\nclients]}$ shared by the set of clients in $\commonconn$, the federator receives $\bscolluders+\bsaddconn$ evaluations of the polynomial $\sum_{\clientindex \in \commonconn} \clientpoly$ which can be interpolated to obtain the coefficients $\sum_{\clientindex \in \commonconn} (\submodel[j] + \subrandvec[j])$ for all $j \in [\bsaddconn]$. Consequently, the federator is always guaranteed to obtain the sum of all padded gradients, i.e., 
    $\sum_{\clientindex \in [\nclients]} (\model[\clientindex] + \randvec[\clientindex])$,
by interpolating the individual polynomials and concatenating the private coefficients.

\paragraph{Secret Key Aggregation}
The result obtained by the federator is statistically independent of the individual gradients and of the sum of gradients $\sum_{\clientindex\in[\nclients]}\model[\clientindex]$ as the individual vectors $\randvec$ are independently and uniformly distributed over $\mathbb{F}_q^\inputdim$. For stronger privacy guarantees, we allow the federator only to obtain the sum of \emph{all} gradients and not the sum of subsets of the gradients. %
To that end, we propose a protocol that only reveals $\sum_{\clientindex \in \nclients} \randvec[\clientindex]$ to the federator. Note that the federator has access to some individual $\model[\clientindex]+\randvec[\clientindex]$. By revealing only the sum of all $\randvec[\clientindex]$'s the federator cannot obtain any information about the sum of any {proper} subset of the clients' models $\model[\clientindex]$.

Each client $\clientindex$ sends their random vector $\randvec[\clientindex]$ to one base station $\clusterindex \in \bsclient$, aggregating all received random vectors. Most likely, each base station $\clusterindex$ owns a partial aggregation of the random vectors of connected clients, denoted by $\randvec[\Sigma]^\clusterindex$. Starting from base station $\clusterindex=1$, the partial aggregation is transmitted to base station $\clusterindex=2$, which sums its partial aggregation to the received partial aggregation and forwards the sum to base station $\clusterindex = 3$ and so on. Hence, base station $\clusterindex>1$ obtains
    $\sum_{j \in [\clusterindex]} \randvec[\Sigma]^j.$
The last base station $\clusterindex = \nclusters$ forwards the aggregate to the federator. Base stations not receiving random vectors can be skipped in the aggregation step. %

\begin{figure}[t]
\centering
    \resizebox{0.6\linewidth}{!}{\input{resources/system_model_simple_example}}
    \caption{Example of wireless architecture for federated learning without relays, where UEs are connected to the federator via base stations. Each UE is connected to at least $\bscolluders+1$ base stations, with $\bscolluders=1$. Solid black links describe aggregated shares. We only depict shares corresponding to evaluation point $\alpha_1$, represented by the leftmost lines of each client. The other two lines describe the shares corresponding to evaluation points $\alpha_2$ and $\alpha_3$ (for client $1$ and $2$). Shares of UEs $1$ and $2$ are aggregated by the three leftmost base stations; shares of UE $3$ are not aggregated. The base stations aggregate the keys, so that the federator cannot decode partial sums of the $g_i$'s.
    }
    \label{fig:systemmodel_simple_example} \vspace{-0.5cm}
\end{figure}

\extstart We provide in \cref{fig:systemmodel_simple_example} an example for the proposed scheme in a wireless system architecture, where two users are connected to the same three base stations and one user is connected to two base stations. For $\bscolluders=1$, the size of the shares sent by UEs $1$ and $2$ are half the size of those of UE $3$. By following the protocol, only the federator can decode the sum $\sum_{\clientindex=1}^3 \model[\clientindex]$. \extend

\subsection{Privacy Analysis}

The introduced scheme guarantees information-theoretic privacy of the clients' data for the cases (a) and (b), which we formally state and prove in \cref{thm:IT-private_scheme}. %

\begin{theorem}[Privacy \extstart of the Scheme\extend]\label{thm:IT-private_scheme}
The aggregation scheme introduced above fulfills the privacy guarantees
\begin{align*}
    &\mathrm{I}\left(\observations^{\mathcal{C}}, \observations^{\mathcal{B}}; \privatedata^{\noncuriousset} \given \extstart\aggregatedgradient, \extend \privatedata^{\uecolluderset} \right) = \mathrm{I}\left(\observations^F, \observations^{\mathcal{C}}; \privatedata^{\noncuriousset} \given \aggregatedgradient, \privatedata^{\uecolluderset} \right) = 0.
\end{align*}
\end{theorem}
\extstart We defer the proof to~\cref{app:proof_of_thm_IT-private_scheme}. \extend

\begin{remark} \label{remark:collusions}
To extend the scheme to the full collusion assumption, specific partial aggregations of the random vectors $\randvec$ have to be made private against any number $\bscolluders$ \extstart of \extend base stations, which can be ensured by a careful instance-dependent system design as detailed in \cref{subsec:arbitrary_colluders}.
\end{remark}

\subsection{Analysis of Communication Cost}

In \cref{thm:IT-private_scheme_communication}, we provide an upper bound on the communication cost $\com_1$ of our scheme and show that $\com_1$ is an increasing function of the privacy parameter $\bscolluders$. However, the bound is independent of $\uecolluders$ since the savings on communication happen through aggregations at the base stations. The exact communication cost depends on the connectivity sets of the clients and on the number of partial aggregations they facilitate. %

\begin{theorem}[Communication Cost with Partial Collusion]\label{thm:IT-private_scheme_communication}
The communication cost $\com_1$ of the scheme in \cref{sec:it_priv} is bounded as
\begin{align}
    \!\! \graddim \bigg(\! \nclients + \nclusters +  \frac{(\nclients\!+\!1) \nclusters}{\nclusters-\bscolluders} \bigg) \! \leq \com_1 \leq \graddim \left( \nclients + \nclusters + 2\nclients(\bscolluders+1) \right)\!. \! \label{eq:loose_bound}
\end{align}
Given some knowledge about the clients' connectivity patterns $\bsclient$, we obtain a tighter bound on the communication cost. Let $\frac{\nclients+1}{\nclients} \leq \beta \leq 2$ be a network-dependent parameter that depends on the connectivity of the clients to the base stations, then the scheme requires a communication cost given by
\begin{align}
    \!\! \com_1  = \graddim  \bigg( \! \nclients \! + \! \nclusters \! + \! \beta \sum_{\clientindex=1 }^\nclients \frac{\bscolluders+\bsaddconn}{\bsaddconn} \! \bigg) \!\! < \!\! \bigg(\! 3 \! + \! \frac{\nclusters-\bscolluders}{\nclients+1} \! \bigg) \mincomm. \! \label{eq:tight_bound}
\end{align}
\end{theorem}
\extstart We defer the proof to~\cref{app:proof_of_thm_IT-private_scheme_communication}. \extend

\begin{remark}
    The communication cost generally decreases with an increasing number of base stations due to an increase of the $\bsaddconn$'s. The communication cost from the base stations to the federator can be decreased by increasing the communication cost from some of the UEs to the base stations. Clients can choose to construct their secret sharing only for a subset of the connected base stations $\bsclient$ to facilitate more partial aggregation opportunities at the base stations by thoroughly aligning their secret shares with a larger set of other clients.
\end{remark}

\begin{figure}[t]
    \centering
    \resizebox{.7\linewidth}{!}{\begin{tikzpicture}

\definecolor{darkgray176}{RGB}{176,176,176}
\definecolor{darkorange25512714}{RGB}{255,127,14}
\definecolor{lightgray204}{RGB}{204,204,204}
\definecolor{steelblue31119180}{RGB}{31,119,180}

\begin{axis}[
height=\figheight,
legend cell align={left},
legend style={fill opacity=0.8, draw opacity=1, text opacity=1, draw=lightgray204},
tick align=outside,
tick pos=left,
width=\figwidth,
x grid style={darkgray176},
xlabel={\(\displaystyle \nu_i\)},
xmin=-0.2, xmax=26.2,
xtick style={color=black},
y grid style={darkgray176},
ylabel={Relative communication \(\displaystyle \frac{\com}{\nclients \graddim}\)},
ymin=0.736, ymax=9.404,
ytick style={color=black}
]
\path [fill=steelblue31119180, fill opacity=0.1]
(axis cs:1,9.01)
--(axis cs:1,5.0104)
--(axis cs:2,3.51025)
--(axis cs:3,3.0102)
--(axis cs:4,2.760175)
--(axis cs:5,2.61016)
--(axis cs:6,2.51015)
--(axis cs:7,2.43871428571429)
--(axis cs:8,2.3851375)
--(axis cs:9,2.34346666666667)
--(axis cs:10,2.31013)
--(axis cs:11,2.28285454545455)
--(axis cs:12,2.260125)
--(axis cs:13,2.24089230769231)
--(axis cs:14,2.22440714285714)
--(axis cs:15,2.21012)
--(axis cs:16,2.19761875)
--(axis cs:17,2.18658823529412)
--(axis cs:18,2.17678333333333)
--(axis cs:19,2.16801052631579)
--(axis cs:20,2.160115)
--(axis cs:21,2.15297142857143)
--(axis cs:22,2.14647727272727)
--(axis cs:23,2.14054782608696)
--(axis cs:24,2.1351125)
--(axis cs:25,2.130112)
--(axis cs:25,3.25)
--(axis cs:25,3.25)
--(axis cs:24,3.26)
--(axis cs:23,3.27086956521739)
--(axis cs:22,3.28272727272727)
--(axis cs:21,3.29571428571429)
--(axis cs:20,3.31)
--(axis cs:19,3.32578947368421)
--(axis cs:18,3.34333333333333)
--(axis cs:17,3.36294117647059)
--(axis cs:16,3.385)
--(axis cs:15,3.41)
--(axis cs:14,3.43857142857143)
--(axis cs:13,3.47153846153846)
--(axis cs:12,3.51)
--(axis cs:11,3.55545454545455)
--(axis cs:10,3.61)
--(axis cs:9,3.67666666666667)
--(axis cs:8,3.76)
--(axis cs:7,3.86714285714286)
--(axis cs:6,4.01)
--(axis cs:5,4.21)
--(axis cs:4,4.51)
--(axis cs:3,5.01)
--(axis cs:2,6.01)
--(axis cs:1,9.01)
--cycle;

\addplot [semithick, steelblue31119180]
table {%
1 9.01
2 6.01
3 5.01
4 4.51
5 4.21
6 4.01
7 3.86714285714286
8 3.76
9 3.67666666666667
10 3.61
11 3.55545454545455
12 3.51
13 3.47153846153846
14 3.43857142857143
15 3.41
16 3.385
17 3.36294117647059
18 3.34333333333333
19 3.32578947368421
20 3.31
21 3.29571428571429
22 3.28272727272727
23 3.27086956521739
24 3.26
25 3.25
};
\addlegendentry{Upper bound on $\com_1$ with $\beta = 2$}
\addplot [semithick, steelblue31119180, dashed]
table {%
1 5.0104
2 3.51025
3 3.0102
4 2.760175
5 2.61016
6 2.51015
7 2.43871428571429
8 2.3851375
9 2.34346666666667
10 2.31013
11 2.28285454545455
12 2.260125
13 2.24089230769231
14 2.22440714285714
15 2.21012
16 2.19761875
17 2.18658823529412
18 2.17678333333333
19 2.16801052631579
20 2.160115
21 2.15297142857143
22 2.14647727272727
23 2.14054782608696
24 2.1351125
25 2.130112
};
\addlegendentry{Lower bound on $\com_1$ with $\beta = \frac{\nclients+1}{\nclients}$}
\addplot [semithick, darkorange25512714]
table {%
1 4.01
2 2.51
3 2.01
4 1.76
5 1.61
6 1.51
7 1.43857142857143
8 1.385
9 1.34333333333333
10 1.31
11 1.28272727272727
12 1.26
13 1.24076923076923
14 1.22428571428571
15 1.21
16 1.1975
17 1.18647058823529
18 1.17666666666667
19 1.16789473684211
20 1.16
21 1.15285714285714
22 1.14636363636364
23 1.1404347826087
24 1.135
25 1.13
};
\addlegendentry{Lower bound $\mincomm$}
\end{axis}

\end{tikzpicture}}
    \caption{Communication cost of our scheme normalized by $\nclients \graddim$ as a function of $\bsaddconn$ for $\nclients=10^4$, $\nclusters=100$ and $\bscolluders = 3$.
    \vspace{-.3cm}}
    \label{fig:commcost}
\end{figure}

\subsection{Numerical \extstart Evaluation \extend}

We plot in \cref{fig:commcost} the communication cost of our scheme as a function of $\bsaddconn$, assuming that $\bsaddconn$ is the same for all clients, for $\nclients=10^4$ clients, $\nclusters=100$ base stations, a gradient dimension of $\graddim=10^6$, and $\bscolluders=3$ colluding base stations\footnote{In the edge case where all clients collude, i.e., $\uecolluders=\nclients$, the privacy guarantee in \eqref{eq:it_privacy} degenerates but still holds.}.

In addition to the constant multiplicative gap between $\com_1$ and $\mincomm$, one can observe a sharp decrease in the communication cost with increasing connectivity $\bsaddconn$ for all $i \in [\nclients]$.

\section{Private Aggregation with Full Collusion} \label{subsec:arbitrary_colluders}

We describe how to extend the proposed scheme so that the federator can collude with any set of $\bscolluders$ base stations, i.e., $\uecolluders=0$, without violating the clients' privacy. We then refine the established methods to account for the full collusion assumption. The core idea is to replace the naive key aggregation with a carefully designed secure multi-party computation that allows for partial aggregations.
More specifically, this requires constructing the secret sharing in \eqref{eq:clientpoly} only for a subset $\bsmain$ of the base stations in $\bsclient$, i.e., $\bsmain \subseteq \bsclient$. In addition, each client secret shares the keys $\randvec$ to another subset of the base stations $\bssec \subseteq \bsclient$. Note that it is not necessary that $\bssec \cap \bsmain = \emptyset$ nor that $\bssec \cup \bsmain = \bsclient$. Let $\bsaddconmain \defeq \vert \bsmain \vert - \bscolluders$ and $\bsaddconsec \defeq \vert \bssec \vert - \bscolluders$, then each of the clients constructs two secret sharings formed by the polynomials
\begin{align*}
    \clientpolygrads &= \sum_{j \in [\bsaddconmain]} x^{j-1} (\submodel[j] + \subrandvec[j]) + \sum_{j \in [\bscolluders]} x^{\bsaddconmain+j-1} \secrandgrads[j], \\
    \clientpolykeys &= \sum_{j \in [\bsaddconsec]} x^{j-1} \subrandvec[j] + \sum_{j \in [\bscolluders]} x^{\bsaddconsec+j-1} \secrandkeys[j], %
\end{align*}
where $\secrandgrads[j]$ and $\secrandkeys[j]$ are vectors chosen \extstart independently and \extend uniformly at random from $\mathbb{F}_q^{\graddim/\bsaddconmain}$ and $\mathbb{F}_q^{\graddim/\bsaddconsec}$, respectively. 
Each client then sends evaluations of $\clientpolygrads$ at distinct points $\alpha_\clusterindex$ to base stations $\clusterindex \in \bsmain$ and evaluations of $\clientpolykeys$ at distinct points $\alpha_\clusterindex^\prime$ to base stations $\clusterindex \in \bssec$. Let $\mainsets \defeq \{\bsmain\}_{\clientindex \in [\nclients]}$ be the distinct sets of base stations used for secretly sharing the padded gradients and $\secsets \defeq \{\bssec\}_{\clientindex \in [\nclients]}$ the distinct sets of base stations used for sharing the keys. 
For every possible set of base stations $\mathcal{Y} \in \mainsets \subset 2^{[\nclusters]}$, let $\commonmain$ be the set of all clients that share the same set $\bsmain = \tmpmain$. Similarly, for $\tmpsec \in \secsets \subset 2^{[\nclusters]}$ let $\commonsec$ be the set of all clients that share the same set $\bssec = \tmpsec$. Every base station $\clusterindex \in \tmpmain$ that receives secret shares from clients $\clientindex \in \commonmain$ aggregates their shares and forwards the aggregations to the federator. The same holds for base station $\clusterindex \in \tmpsec$ and clients $\clientindex \in \commonsec$.

The federator receives the aggregated shares of padded gradients of clients $\{\commonmain\}_{\tmpmain \in \mainsets}$ and the aggregated shares of keys of clients $\{\commonsec\}_{\tmpsec \in \secsets}$. To guarantee privacy, we have to ensure that the (unified) power sets of these two sets only intersect in $\mathcal{N}$, which is required for the federator to reconstruct the sum of the clients' gradients, i.e., 
\begin{align}
    \!\!\! \big\{ \! \textstyle\bigcup \tmpset \! \mid \!  \tmpset \! \in 2^{\{\commonmain\}_{\tmpmain \in \mainsets}}\big \} \cap \big\{ \! \textstyle\bigcup \tmpset \! \mid \! \tmpset \! \in 2^{\{\commonsec\}_{\tmpsec \in \secsets}}  \big\} = \mathcal{N}, \label{eq:privacy_condition}
\end{align}
\extstart where $\bigcup \tmpset$ is the union of all sets contained in $\tmpset$. \extend

However, privacy could break if only one client colluded with the other parties, i.e., $\uecolluders = 1$. \extstart For example, if for two chosen sets $\tmpmain \in \mainsets$ and $\tmpsec \in \secsets$ client $\clientindex$ is the only difference, i.e., $\tmpsec \setminus \tmpmain = \{\clientindex\}$, then a collusion of client $\clientindex$ with the federator reveals the sum of gradients of all clients in $\tmpmain$. \extend In fact, for any number $\uecolluders > 0$, the two sets have to fulfill some notion of distance requirements, which we describe using a coding-theoretic approach. Consider the generator matrix $\mathbf{G}_1 \in \mathbb{F}_2^{\vert \mainsets \vert \times \nclients}$ of a linear block code, where each row $j \in [\vert \mainsets \vert]$ represents a set of base stations in $\mainsets$ and each column represents a client. The $\extstart(j, \clientindex)\extend$-th entry of $\mathbf{G}_1$ is set to $1$ if client $\clientindex$ sends shares of its padded gradient to the set of base stations in $\mainsets$ indexed by $j$ and $0$ otherwise. The code $\mathcal{C}_1$ generated by $\mathbf{u} \mathbf{G}_1$ for all $\mathbf{u} \in \mathbb{F}_2^{\vert \mainsets \vert}$ spans the space earlier described by $\big\{ \textstyle\bigcup \tmpset \mid  \tmpset \in 2^{\{\commonmain\}_{\tmpmain \in \mainsets}}\big\}$. By construction, the weight of each column in $\mathbf{G}_1$ is exactly one. Thus, the sum of rows of $\mathbf{G}_1$, i.e., $\mathbf{1} \cdot \mathbf{G}_1$ represents the sum of all padded gradients $\sum_{\clientindex=1}^\nclients \model + \randvec$.

Similarly, for describing the key aggregation process, we define a generator matrix $\mathbf{G}_2 \in \mathbb{F}_2^{\vert \secsets \vert \times \nclients}$, where each row $j \in [\vert \secsets \vert]$ represents a set of base stations in $\secsets$. Hence, the $\extstart (j, \clientindex)\extend$-th entry of $\mathbf{G}_2$ equals one if client $\clientindex$ sends shares of its key to the $j$-th \extstart set of base stations \extend in $\secsets$. The code $\mathcal{C}_2$ generated by $\mathbf{u} \mathbf{G}_2$ for $\mathbf{u} \in \mathbb{F}_2^{\vert \secsets \vert}$ spans the set described by $\big\{ \textstyle\bigcup \tmpset \mid  \tmpset \in 2^{\{\commonsec\}_{\tmpsec \in \secsets}}  \big\}$.

For the case of $\uecolluders = 0$, the privacy condition in \eqref{eq:privacy_condition} translates to finding generators $\mathbf{G}_1$ and $\mathbf{G}_2$ such that the minimum distance $d_{\min}^{\mathcal{C}_1, \mathcal{C}_2}$ between the two codes $\mathcal{C}_1$ and $\mathcal{C}_2$ defined next, is at least one, i.e., $d_{\min}^{\mathcal{C}_1, \mathcal{C}_2}\geq 1$, with $$\displaystyle d_{\min}^{\mathcal{C}_1, \mathcal{C}_2} \defeq \!\! \min_{%
\substack{\mathbf{c_1} \in \mathcal{C}_1, \mathbf{c_2} \in \mathcal{C}_2,\\ \text{if }\mathbf{c}_1 \in \{\mathbf{0}, \mathbf{1}\}, \text{ then }\mathbf{c}_1 \neq \mathbf{c}_2}}\!\!
\text{wt}(\mathbf{c_1} - \mathbf{c_2}),$$%
where $\text{wt}(\cdot)$ is the 
Hamming weight, and $\mathbf{0}$ and $\mathbf{1}$ are the all-zero and all-one vectors, respectively. 
In other words, all rows of $\mathbf{G}_1$ are linearly independent from all rows of $\mathbf{G}_2$ when excluding the all-one and all-zero rows.

Thus, privacy against any number of colluding parties $\uecolluders > 0$ is guaranteed by ensuring $d_{\min}^{\mathcal{C}_1, \mathcal{C}_2} \geq 1+ \uecolluders$. %
This is equivalent to saying that even after removing any $\uecolluders$ columns from both $\mathbf{G}_1$ and $\mathbf{G}_2$ (same columns are removed), the rows of the submatrix of $\mathbf{G}_1$ (with the remaining columns) are linearly independent from the rows of the submatrix of $\mathbf{G}_2$ (with the remaining columns) when excluding the all-one and all-zero rows. Translated to our private gradient and key aggregation process, despite knowing all partial aggregations of padded gradients and keys in addition to the keys and gradients of any $\uecolluders$ clients, it is impossible to recover any partial aggregation of plain gradients. This holds since there will always be at least one unknown key $\randvec$ with maximum entropy remaining in every partial sum of $\sum_{\clientindex \in \extstart \cup \extend \tmpset} \model + \randvec$ for some $\tmpset$ that makes it impossible to recover $\sum_{\clientindex \in \extstart \cup \extend \tmpset} \model$.

The problem boils down to finding sets $\bsmain \subseteq \bsclient$ and $\bssec \subseteq \bsclient$ for all clients $\clientindex \in [\nclients]$ which generate suitable sets $\mainsets$, $\secsets$ and generator matrices $\mathbf{G}_1$ and $\mathbf{G}_2$ with the desired properties, i.e., the weight of each column in $\mathbf{G_1}$ and $\mathbf{G}_2$ is exactly one, the rows of both matrices are linearly independent, and the minimum distance between the codes $\mathcal{C}_1$ and $\mathcal{C}_2$ satisfies $d_{\min}^{\mathcal{C}_1, \mathcal{C}_2} \geq 1+ \uecolluders$. Since the all-one vector is in the span of both generator matrices, this directly implies that rows of $\mathbf{G}_1$ and $\mathbf{G}_2$ must have weight at least $1 + \uecolluders$, and hence that $d_{\min}^{\mathcal{C}_1} \geq 1 + \uecolluders$ and $d_{\min}^{\mathcal{C}_2} \geq 1 + \uecolluders$. This holds because the weight of every column in $\mathbf{G}_1$ and $\mathbf{G}_2$ is exactly one, therefore linear combinations of rows in $\mathbf{G}_1$ and $\mathbf{G}_2$ can only increase the weight. Hence, the minimum distance is the minimum weight among the rows.

Finding suitable sets $\bsmain$ and $\bssec$ as described above results in a scheme whose communication cost can be described as in \cref{thm:communication_relayed}.
\begin{theorem}[Communication Cost with Full Collusion] \label{thm:communication_relayed}
Let $\frac{\nclients + 1}{\nclients} \leq \gamma \leq \frac{2 + \uecolluders}{1 + \uecolluders}$ be a network-dependent parameter determined by the row weights of $\mathbf{G}_1$ and $\mathbf{G}_2$. Then, the communication cost of the scheme with full collusion is
    \begin{align*}
    \com_2 &= \gamma \cdot \graddim \left(\sum_{\clientindex = 1}^\nclients \frac{\bsaddconmain + \bscolluders}{\bsaddconmain} + \frac{\bsaddconsec + \bscolluders}{\bsaddconsec}\right) \\
    &= 2 \gamma \cdot \xi \cdot \graddim \sum_{\clientindex = 1}^\nclients \extstart \frac{\bsaddconn + \bscolluders}{\bsaddconn} \extend,
    \end{align*}
    where $\xi$ with $1 \leq \xi \leq \nclusters - \bscolluders$ describes the interrelation between $\bsaddconn$, $\bsaddconmain$ and $\bsaddconsec$.
\end{theorem}

Before proving the theorem, we describe the intuition of the results. The first equation in the theorem highlights that the communication cost decreases as $\bsaddconmain$ or $\bsaddconsec$ increases. Compared to \cref{thm:IT-private_scheme_communication}, the upper bound on the communication cost now depends on the clients' collusion parameter $\uecolluders$, which is reflected by $\gamma$. When $\uecolluders$ increases, the maximum communication cost decreases as the minimum number of partial aggregations (reflected in $d_{\min}^{\mathcal{C}_1,\mathcal{C}_2}$) required to meet the privacy condition must increase. However, the larger $\uecolluders$, the more connections from clients to base stations and relays are required, which imposes more constraints on constructing such a private aggregation scheme. The second equation relates the communication cost to the connectivity of each client, which can be used for comparability with previous results.

\begin{proof}[Proof of \cref{thm:communication_relayed}]
We describe the cost of communication by summing the cost of secret sharing over all sets $\bsmain$, which is equivalent to weighting the sets in $\mainsets$ by the number of clients that share this set, i.e., by $\vert \commonmain \vert$. For each set in $\mainsets$ the base stations forward the aggregated secret sharings to the federator. The same argument applies to the secret sharings of the keys. Hence, we get
    \begin{align*}
    \com_2 &= \sum_{\tmpmain \in \mainsets} (\vert \commonmain \vert + 1) \frac{\graddim \vert \tmpmain \vert}{\vert \tmpmain \vert - \bscolluders} + \sum_{\tmpsec \in \secsets} (\vert \commonsec \vert + 1) \frac{\graddim \vert \tmpsec \vert}{\vert \tmpsec \vert - \bscolluders} \\
    &= \sum_{\clientindex = 1}^\nclients \frac{\graddim \vert \bsmain \vert}{\vert \bsmain \vert - \bscolluders} \left(1 + \frac{1}{n_{\vert \bsmain \vert}}\right) + \frac{\graddim \vert \bssec \vert}{\vert \bssec \vert - \bscolluders} \left(1 + \frac{1}{n_{\vert \bssec \vert}}\right),
\end{align*}
where $\nclients_{\vert \tmpmain \vert} \defeq \vert \commonmain \vert$ describes the number of clients that share $\tmpmain$, i.e.,  the latter expression is equivalent since $\sum_{\clientindex = 1}^\nclients \frac{\graddim \vert \bsmain \vert}{\vert \bsmain \vert - \bscolluders} = \sum_{\tmpmain \in \mainsets} \vert \commonmain \vert \frac{\graddim \vert \tmpmain \vert}{\vert \tmpmain \vert - \bscolluders}$ and $\sum_{\clientindex = 1}^\nclients \frac{\graddim \vert \bsmain \vert}{\vert \bsmain \vert - \bscolluders} \frac{1}{n_{\vert \bsmain \vert}} = \sum_{\tmpmain \in \mainsets} \frac{\graddim \vert \tmpmain \vert}{\vert \tmpmain \vert - \bscolluders}$.
To bound $\nclients_{\vert \bsmain \vert}$ and $\nclients_{\vert \bssec \vert}$, we consider the best-case and worst-case scenarios where either all or none of the clients' shares can be aggregated. By bounding $1 + \uecolluders \leq \nclients_{\vert \bsmain \vert}, \nclients_{\vert \bssec \vert} \leq \nclients$ for any $\clientindex$ and hence choosing $\frac{\nclients + 1}{\nclients} \leq \gamma \leq \frac{2 + \uecolluders}{1 + \uecolluders}$ we obtain the theorem statement. We continue to relate the communication cost to the clients' connectivity parameter $\bsaddconn$'s. Since the sets $\bsmain$ and $\bssec$ have at most the size of $\bsclient$, it trivially holds that
\begin{align}\label{eq:lbc2proof}
    \com_2 &\geq 2 \gamma \sum_{\clientindex = 1}^\nclients \frac{\graddim \vert \bsclient \vert}{\vert \bsclient \vert - \bscolluders}.
\end{align}
Further, we can derive the following upper bound with respect to the $\bsclient$'s.
\begin{align}
    \com_2 
    &= \gamma \sum_{\clientindex = 1}^\nclients \left(\frac{\vert \bsmain \vert (\vert \bsclient \vert - \bscolluders)}{\vert \bsclient \vert (\vert \bsmain \vert - \bscolluders)} + \frac{\vert \bssec \vert (\vert \bsclient \vert - \bscolluders)}{\vert \bsclient \vert (\vert \bssec \vert - \bscolluders)}\right) \frac{\graddim \vert \bsclient \vert}{\vert \bsclient \vert - \bscolluders} \nonumber\\
    &\leq \gamma \sum_{\clientindex = 1}^\nclients \left(\frac{\bsaddconn}{\bsaddconmain} + \frac{\bsaddconn}{\bsaddconsec} \right) \frac{\graddim \vert \bsclient \vert}{\vert \bsclient \vert - \bscolluders} \nonumber\\
    &\leq \gamma \max_{\clientindex \in [\nclients]} \left(\frac{\bsaddconn}{\bsaddconmain} + \frac{\bsaddconn}{\bsaddconsec} \right) \sum_{\clientindex = 1}^\nclients \frac{\graddim \vert \bsclient \vert}{\vert \bsclient \vert - \bscolluders} \nonumber\\
    &\leq 2 \gamma (\nclusters - \bscolluders) \sum_{\clientindex = 1}^\nclients \frac{\graddim \vert \bsclient \vert}{\vert \bsclient \vert - \bscolluders},\label{eq:ubc2proof}
\end{align}
where the latter inequality follows since $\bsaddconmain, \bsaddconsec \geq 1$ and $\bsaddconn \leq \nclusters - \bscolluders$. Combining~\eqref{eq:lbc2proof} and~\eqref{eq:ubc2proof}, there exists a $\xi$ with $1 \leq \xi \leq \nclusters - \bscolluders$ such that the statement of the theorem holds.
\end{proof}

In \cref{thm:IT-private_scheme_seckey}, we prove the privacy of the proposed extension under the assumption of full collusion.

\begin{theorem}[Privacy with Full Collusion]\label{thm:IT-private_scheme_seckey}
The extended aggregation scheme introduced in \cref{subsec:arbitrary_colluders} with secure key aggregation fulfills the privacy guarantee
\begin{align*}
    &\mathrm{I}\left(\observations^F, \observations^\mathcal{B}, \observations^{\mathcal{C}}; \privatedata^{\noncuriousset} \vert \aggregatedgradient, \privatedata^{\uecolluderset} \right) = 0.
\end{align*}
\end{theorem}
\extstart We defer the proof to~\cref{app:proof_of_thm_IT-private_scheme_seckey}. \extend

The following example illustrates the above extension of the private aggregation scheme to full collusion and compares the communication costs of both schemes to the lower bound.
\begin{example}

Consider a setting with $\nclients=6$ clients and $\nclusters=5$ base stations, where one client is allowed to collude with two base stations and the federator, i.e., $\uecolluders=1$ and $\bscolluders=2$. The clients have the following connectivity sets
\begin{align*}
    \bsclient[1] = \bsclient[2] &= \{1, 2, 3, 5\}, &&
    \bsclient[3] = \{1, 2, 3, 4, 5\}, \\
    \bsclient[4] &= \{2, 3, 4, 5\},
    && \bsclient[5] = \{1, 2, 4, 5\}, \\
    \bsclient[6] &= \{1, 2, 5\}.
\end{align*}

The \extstart distinct sets $\tmpmain_j^\prime$ and $\tmpsec_j^\prime$ are chosen to be $\tmpmain_1^\prime = \{1, 3, 5\}$, $\tmpmain_2^\prime = \{2, 3, 4, 5\}$, $\tmpmain_3^\prime = \{1, 2, 5\}$, $\tmpsec_1^\prime = \{1, 2, 3, 5\}$, $\tmpsec_2^\prime = \{2, 4, 5\}$ and $\tmpsec_3^\prime = \{1, 2, 5\}$.
In addition, we construct the following two generator matrices $\mathbf{G}_1$ and $\mathbf{G}_2$ such that $d_{\min}^{\mathcal{C}_1, \mathcal{C}_2} \geq 2$, and $\tmpmain_j^\prime = \bsmain \subseteq \bsclient$ for all clients $\clientindex$ in the support of the $j$-th row of $\mathbf{G}_1$ and $\tmpsec_j^\prime = \bssec \subseteq \bsclient$ \extend for all clients $\clientindex$ in the support of the $j$-th row of $\mathbf{G}_2$. The matrices $\mathbf{G}_1$ and $\mathbf{G}_2$ read as
\begin{align*}
    \mathbf{G}_1 &= \begin{pmatrix}
    1 & 1 & 0 & 0 & 0 & 0 \\
    0 & 0 & 1 & 1 & 0 & 0 \\
    0 & 0 & 0 & 0 & 1 & 1 
    \end{pmatrix},& \!\! \mathbf{G}_2 &= \begin{pmatrix}
    0 & 1 & 1 & 0 & 0 & 0 \\
    0 & 0 & 0 & 1 & 1 & 0 \\
    1 & 0 & 0 & 0 & 0 & 1 
    \end{pmatrix}.
\end{align*}

This construction attains the privacy guarantee with full collusion according to \cref{thm:IT-private_scheme_seckey} and achieves a communication cost of $\com_2 = 48 \graddim$. Under partial collusion, we achieve a communication cost of $\com_1 = 23.33 \graddim$ as compared to the lower bound, which is $\mincomm = 15.67 \graddim$.

\end{example}

\section{Private Aggregation for Wireless Systems With Relays} \label{sec:extended_scheme}

In the following, we consider the generalized multi-hierarchical system architecture, where the base stations relay information to the federator through other nodes on the wired link. This system models many practical common infrastructures where the federator cannot be directly connected to the base stations. For clarity of presentation, we restrict ourselves to investigating one additional level of hierarchy, whose entities we refer to as \emph{relay nodes}. The total number of relay nodes is referred to as $\ntotalrelays$. We impose certain privacy conditions on each layer, i.e., how many entities $\relaycollusions$ are allowed to collude without leaking any private information. Introducing $\mathcal{R}$ as the set of all $\relaycollusions$ colluding relay nodes, then the information-theoretic privacy requirements introduced in \cref{sec:system_architecture} formally translates to the following privacy notion with full collusion:
\begin{align}
    &\mathrm{I}\left(\observations^{\mathcal{C}}, \observations^\mathcal{B}, \observations^{\mathcal{R}}, \observations^F; \privatedata^{\noncuriousset} \given \aggregatedgradient, \privatedata^{\uecolluderset} \right) = 0. \label{eq:strong_privacy2}
\end{align}
Extending our scheme from \cref{sec:it_priv} to this more general setting requires carefully selecting base stations and relay nodes together with their evaluation points. Constrained with the privacy guarantee, our target is to minimize the communication cost. We only explicitly present a scheme that attains privacy conditions with partial collusion described by:
\begin{align}
    &\mathrm{I}\left(\observations^{\mathcal{C}}, \observations^{\mathcal{B}}; \privatedata^{\noncuriousset} \given \extstart\aggregatedgradient,\extend \privatedata^{\uecolluderset} \right) = 0 \text{ and } \label{eq:it_privacy2}\\
    &\mathrm{I}\left(\observations^{\mathcal{C}}, \observations^{\mathcal{R}}, \observations^F; \privatedata^{\noncuriousset} \given \extstart\aggregatedgradient,\extend \privatedata^{\uecolluderset} \right) = 0 \label{eq:it_privacy_federator2}
\end{align}

However, similar to \cref{subsec:arbitrary_colluders}, our scheme can be modified to provide privacy under full collusion as in \eqref{eq:strong_privacy2}, which we detail at the end of this section.
To attain the privacy level imposed by \eqref{eq:it_privacy2} and \eqref{eq:it_privacy_federator2}, it is fundamentally required that for each client $\clientindex$ there exist at least $\bscolluders + 1$ base stations in turn connected to $\relaycollusions + 1$ distinct relay nodes.

Each client $\clientindex$ determines a certain set of base stations $\bsset$ in reach, which are capable of relaying the received shares through a sufficient set of relay nodes $\relayset$ such that $\bscolluders + 1 \leq \vert \bsset \vert \leq \nclusters$ and $\relaycollusions + 1\leq \vert \relayset \vert \leq \ntotalrelays $. This is described by the sets of relays $\relaybs$ to which each base station $\clusterindex$ is connected. Clients with the same $\bsset$ should choose\footnote{This assumption is because shares of clients with the same $\bsset$ will get aggregated and hence forwarded to the same relays by system design.} the same $\relayset$.
For ease of exposition, we will assume that $\vert \bsset \vert = \vert \relayset \vert > \maxcolluders \defeq \max\{\bscolluders, \relaycollusions\}$, which can likely be ensured as the number of available relay nodes can be assumed to exceed the size of $\bsset$. Consequently, let $\maxcard$ be the maximum number of nodes for which the secret sharing of any client can be designed, i.e., $\maxcard \defeq \min\{\nclusters, \ntotalrelays\}$. Further, we define $\cardset \defeq \vert \bsset \vert - \maxcolluders = \vert \relayset \vert - \maxcolluders$.
Base station $\clusterindex \in \bsset$ will relay the share from client $\clientindex$ to relay node $\relayidx_{\clientindex, \clusterindex} \in \relayset \cap \relaybs$. 
We later discuss the implications of the choice of $\bsset$, $\relayset$ and $\{\relayidx_{\clientindex, \clusterindex}\}_{\clusterindex \in \bsset}$ to minimize the communication cost. %

\paragraph{Secret Key Generation and Secret Sharing Construction}
Similar to \cref{sec:it_priv}, each client first generates key $\randvec$ uniformly at random from $\mathbb{F}_q^\graddim$ and splits $\model$ and $\randvec$ into $\cardset$ parts $\submodel$ and $\subrandvec$ for $j \in [\cardset]$. Those vectors are then encoded into a polynomial $\clientpoly$ as in \eqref{eq:clientpoly}. To minimize the communication cost, the evaluation points $\alpha_{\clientindex, \clusterindex}$ assigned to base station $\clusterindex \in \bsset$ should be selected carefully under consideration of $\left\{\bsset, \relayset, \{\relayidx_{\clientindex, \clusterindex}\}_{\clusterindex \in \bsset}\right\}$ for all clients $\clientindex \in [\nclients]$ to maximize partial aggregation opportunities for base stations and relay nodes. \extstart For any two $\clusterindex, \clusterindex^\prime \in \bsset$, the relay nodes must be distinct to avoid that one relay node receives two shares of the same secret sharing, thus potentially violating the privacy guarantee, i.e., $\relayidx_{\clientindex, \clusterindex} \neq \relayidx_{\clientindex, \clusterindex^\prime}$. \extend Aggregations can happen if either i) multiple clients $\clientindex \in \commonbs[\tmpbs]$ share the same set of base stations $\tmpbs$ with common evaluation points for each base station $\clusterindex \in \tmpbs$ thereof, i.e., $\{\alpha_{\clientindex, \clusterindex}\}_{\clientindex \in \commonbs[\tmpbs]}$ contains exactly one element for all $\clusterindex \in \tmpbs$ or ii) multiple clients $\clientindex \in \commonrelay$ share the same set of relay nodes $\tmprelay$ with common evaluation points $\{\alpha_{\clientindex, \clusterindex}\}_{\clientindex \in \commonrelay}$ that are distinct for all $\relayidx_{\clientindex, \clusterindex} \in \tmprelay$ and \extstart where the corresponding polynomial evaluations \extend have been relayed from base stations $\clusterindex \in \tmpbs$. Note that we can express the evaluations at relay node $\relayidx = \relayidx_{\clientindex, \clusterindex}$ as $\alpha_{\clientindex, \relayidx} = \alpha_{\clientindex, \clusterindex}$. %

When choosing $\left\{\bsset, \relayset, \{\relayidx_{\clientindex, \clusterindex}\}_{\clusterindex \in \bsset}\right\}$ for all clients, partial aggregations at base stations should be preferred over relay node aggregations, and large sets $\tmpbs$ or $\tmprelay$ should be prioritized due to small share sizes. For an optimal solution, this combinatorial optimization problem should be treated specifically for a given network architecture. However, we propose the following assignment of evaluation points that allows for a substantial amount of aggregation opportunities. Let $\alpha_1, \dots, \alpha_\maxcard$ be distinct evaluation points. Client $\clientindex$ sends $\vert \bsset \vert$ evaluation points $\alpha_1, \dots, \alpha_{\vert \bsset \vert}$ to the base stations in $\bsset$ in ascending order of their indices. Hence, for each client, there exists a unique mapping from $(\alpha_1, \dots, \alpha_{\vert \bsset \vert})$ to $\alpha_{\clientindex, \clusterindex}$ for all $\clusterindex \in \bsset$. 

\paragraph{Processing the Shares by Base Stations}
If all base stations $\clusterindex \in \tmpbs$ receive shares corresponding to the same evaluation point $\alpha_{\clientindex, \clusterindex}$ from multiple clients $\clientindex \in \commonbs[\tmpbs]$, they first sum the contributions, i.e., $\sum_{\clientindex \in \commonbs[\tmpbs]} \clientpoly[\alpha_{\clientindex, \clusterindex}]$. The results are then forwarded to relay $\relayidx_{\clientindex, \clusterindex}$. Note that $\relayidx_{\clientindex, \clusterindex}$ will be the same for all $\clientindex \in \commonbs[\tmpbs]$. If shares of certain clients cannot be aggregated, the base stations forward $\clientpoly[\alpha_{\clientindex, \clusterindex}]$ to relay $\relayidx_{\clientindex, \clusterindex}$.

\paragraph{Processing of Shares by Relay Nodes}
Relay node $\relayidx$ obtains potentially aggregated shares $\sum_{\clientindex \in \commonbs[\tmpbs]} \clientpoly[\alpha_{\clientindex, \relayidx}]$ for all $\tmpbs \in \{\bsset\}_{\clientindex \in [\nclients]}$ from all base stations \extstart $\clusterindex \in \tmpbs$ \extend where $\relayidx = \relayidx_{\clientindex, \clusterindex}$. Note that $\commonbs[\tmpbs]$ can contain one element; that client's contribution is not aggregated. If all relay nodes $\relayidx \in \tmprelay$ obtain shares $\sum_{\clientindex \in \commonbs[\tmprelay]} \clientpoly[\alpha_{\clientindex, \relayidx}]$ at common evaluation points $\alpha_{\clientindex, \relayidx}$ from multiple clients $\clientindex \in \commonrelay[\tmprelay]$ where $\commonbs[\tmpbs] \subset \commonrelay[\tmprelay]$, it sums the contributions to obtain $\sum_{\clientindex \in \commonrelay[\tmprelay]} \clientpoly[\alpha_{\clientindex, \relayidx}]$ and forwards the result to the federator. By \extstart the proposed \extend design of the evaluation points, \extstart relay nodes can always aggregate the received contributions from clients that share the same set of relays. This holds since the evaluation points are sorted by the indices of the base stations and relays; hence, the evaluations points are aligned whenever clients share the same set of relays. \extend

\paragraph{Interpolation by Federator}
The federator obtains polynomial evaluations $\sum_{\clientindex \in \commonbs[\tmprelay]} \clientpoly[\alpha_{\clientindex, \relayidx}]$ for a sufficient number of $\cardset + \maxcolluders$ evaluations, where $\tmprelay \in \{\relayset\}_{\clientindex \in [\nclients]}$ such that \extstart $ \cup_{\tmprelay \in \{\relayset\}_{\clientindex \in [\nclients]}} \commonbs[\tmprelay] = [\nclients]$ \extend. That is, the sum $\sum_{\clientindex \in [\nclients]} \clientpoly[\alpha_{\clientindex, \relayidx}]$ can be reconstructed and the polynomial can be interpolated.

\paragraph{Secret Key Aggregation}
Equivalent to \cref{sec:it_priv}, it has to be made sure that the federator only obtains the sum of all keys $\sum_{\clientindex = 1}^{\nclients} \randvec$. Therefore, each client selects a base station to which it sends its random key. The base stations then sequentially aggregate the keys and pass the aggregate through any of the relay nodes to the federator.

\begin{figure}[t]
\centering
    \resizebox{0.7\linewidth}{!}{\input{resources/system_model_extended_example}}
    \caption{Example of generalized hierarchical architecture for federated learning, where UEs are connected to the federator via base stations and relay nodes. Each UE is connected to several base stations. One base station is connected to multiple relay nodes. Each relay (R) is connected to the federator. Solid black links describe aggregated shares. We only depict shares corresponding to evaluation point $\alpha_1$, represented by the leftmost lines of each client. The other two lines describe the shares corresponding to evaluation points $\alpha_2$ and $\alpha_3$. Shares of UEs $1$ and $2$ are aggregated by the first three base stations from the left. The shares of UE $3$ are aggregated by the leftmost three relays, whereas the shares of UE $4$ cannot be aggregated. 
    }
    \label{fig:systemmodel_extended} \vspace{-0.5cm}
\end{figure}

We graphically describe an example of the generalized system architecture and the proposed private aggregation scheme in \cref{fig:systemmodel_extended}.
The described scheme is restricted to guaranteeing privacy under partial collusion as defined in \eqref{eq:it_privacy2} and \eqref{eq:it_privacy_federator2}. However, our scheme can be modified to enhance the privacy targeting \eqref{eq:strong_privacy2}. First, one must ensure that the secret sharings resist $\bscolluders$ base stations colluding with $\relaycollusions$ relay nodes. Since base stations forward their received shares through relay nodes, the clients must design their secret shares for $\bscolluders + \relaycollusions$ colluders to account for arbitrary collusion sets. However, since relay nodes provide additional aggregation opportunities, the conditions imposed on the matrices $\mathbf{G}_1$ and $\mathbf{G}_2$ \extstart to fulfil the privacy guarantee in \eqref{eq:privacy_condition} \extend as described in \cref{subsec:arbitrary_colluders} are less restrictive and easier to match \extstart in cases where shares are not directly forwarded from the base stations to the federator, but relayed through relay nodes. This is because relays facilitate additional partial aggregations in addition to those carried out by the base stations, thus enabling to construct codes $\mathcal{C}_1$ and $\mathcal{C}_2$ (as described in \cref{subsec:arbitrary_colluders}) with larger distance $d_{\min}^{\mathcal{C}_1, \mathcal{C}_2}$ since the weights of the codewords can only increase with additional aggregations. \extend %

\subsection{Privacy Analysis}
We state and prove the privacy guarantees achieved by the introduced scheme for a generalized hierarchical network architecture in \cref{thm:IT-private_scheme_extended}.

\begin{theorem}[Privacy with Partial Collusion]\label{thm:IT-private_scheme_extended}
The aggregation scheme introduced in \cref{sec:extended_scheme} fulfills the privacy guarantees
\begin{align*}
    \mathrm{I}\left(\observations^{\mathcal{C}}, \observations^{\mathcal{B}}; \privatedata^{\noncuriousset} \vert \extstart \aggregatedgradient,\extend \privatedata^{\uecolluderset} \right) = \mathrm{I}\left(\observations^F, \observations^{\mathcal{C}}, \ext{\observations^{\mathcal{R}}}; \privatedata^{\noncuriousset} \vert \aggregatedgradient, \privatedata^{\uecolluderset} \right) = 0.
\end{align*}
\end{theorem}
\extstart We defer the proof to~\cref{app:proof_of_thm_IT-private_scheme_extended}. \extend

\subsection{Analysis of Communication Cost}

We first derive a lower bound on the communication cost of any private aggregation scheme satisfying the partial collusion assumption in the generalized hierarchical setting. Then, we continue to analyze the cost of our scheme. For the lower bound, we require the following definitions. Assume that client $\clientindex$ is connected to $\bscolluders < \lbsset \leq \nclusters$ base stations which are in turn connected to the federator through at least $\relaycollusions + 1$ independent paths. Similarly, let client $\clientindex$ be connected to $\relaycollusions < \lrelayset \leq \ntotalrelays$ relay nodes through at least $\bscolluders + 1$ different base stations.

\begin{theorem}[Lower Bound on the Communication Cost] \label{thm:converse2}
Given an architecture as in \cref{sec:extended_scheme} where each client $\clientindex \in [\nclients]$ is characterized by $(\lbsset, \lrelayset)$.
Any scheme that guarantees information-theoretic privacy, for all gradient distributions, against $\uecolluders$ clients colluding with $\bscolluders$ base stations (cf.~\eqref{eq:it_privacy2}) or with $\relaycollusions$ relay nodes and the federator (cf.~\eqref{eq:it_privacy_federator2}) requires a communication cost $\com_\mathrm{R}$ bounded from below by 
\extstart
\begin{align*}
    \com_\mathrm{R} \geq \mincommr \defeq \graddim \left(\max_{\clientindex \in  [\nclients]} \frac{\lrelayset}{\lrelayset - \relaycollusions} + \sum_{\clientindex = 1}^{\nclients} \frac{\lbsset}{\lbsset - \bscolluders} + \right. \nonumber \\
    \left. \max_{\clientindex \in  [\nclients]} \max\left\{\frac{\lbsset}{\lbsset - \bscolluders},\frac{\lrelayset}{\lrelayset - \relaycollusions} \right\} \right). %
\end{align*} \extend
\end{theorem}%
\begin{proof}%
We assume that the clients' gradients are independent and uniformly distributed in $\mathbb{F}_q^\inputdim$, and cannot be compressed. Hence, each client must send a vector of dimension $\graddim$ through base stations and relay nodes. We prove the lower bound on the communication cost by a union-bound argument. 
\extstart
The messages sent from each client $\clientindex$ to its connected base stations must form an $(n=\lbsset, k=\lbsset, z=\bscolluders)$ secret sharing of $\model[\clientindex]$.
  Using~\cref{prop:sharesize} and summing over all clients, we obtain
\extend
\begin{align*}
    \uecom_\mathrm{R} \geq \sum_{\clientindex \in  [\nclients]} \frac{\graddim \lbsset}{\lbsset - \bscolluders}. %
\end{align*}
\extstart
  The communication cost from base stations to relays and from the relays to the federator proceed similarly.
\extend 
Since the shares of all clients could potentially be aggregated at base stations or relay nodes, we consider only the client with the worst connectivity \extstart for the communication from the relays to the federator \extend and, hence, the most expensive secret sharing in terms of share size, i.e.,
\begin{align*}
    \rfcom_\mathrm{R} \geq \max_{\clientindex \in  [\nclients]} \frac{\graddim \lrelayset}{\lrelayset - \extstart \relaycollusions \extend}. %
\end{align*}
Since the secret shares of the clients can potentially be aggregated by base stations, for the least required communication overhead from base stations to relay nodes we have to consider the maximum cost among all clients. More specifically, the lower bound is determined by the secret sharing that provides the worst connectivity over colluding ratio, i.e.,
\begin{align*}
    \bsrcom_\mathrm{R} \geq \max_{\clientindex \in  [\nclients]} \max\left\{\frac{\graddim \lbsset}{\lbsset - \bscolluders},\frac{\graddim \lrelayset}{\lrelayset - \relaycollusions} \right\}. %
\end{align*}
Combining yields $\com_\mathrm{R} \geq \uecom_\mathrm{R} + \bsrcom_\mathrm{R} + \rfcom_\mathrm{R}$ and concludes the proof.
\end{proof}

We continue to analyze the communication cost achieved by the generalized private aggregation scheme. On a high level, the communication cost of our scheme mainly depends on three factors: 1) how many base stations each client sends its shares to, 2) how many clients choose the same set of base stations to send their shares to, and 3) how many clients whose secret shares have not been aggregated by base stations share the same set of relay nodes.

\begin{theorem}[Communication Cost]
Assuming that there exists one set of relay nodes that all clients with the same number $\vert \bsset \vert$ share, the communication cost of the scheme in \cref{sec:extended_scheme} is bounded by
\begin{align*}
    \graddim \bigg(\! \nclients + \nclusters + 1 +  \frac{(\nclients\!+\!2) \maxcard}{\maxcard-\maxcolluders} \bigg) \leq \com_3 \leq &\graddim \big( \nclients (2\maxcolluders + 3) + 2 \nclusters + \\ 
    & (\maxcolluders (H_{\extstart \maxcard - \maxcolluders \extend} - 1)) \big),
\end{align*}
with $H_n$ being the harmonic number. Given some knowledge about the clients' connectivity patterns $\bsset$, we obtain a tighter bound on the communication cost. Let $\frac{\nclients+2}{\nclients} \leq \beta^\prime \leq 3$ and $\tau \defeq \max_{\clientindex \in [\nclients]} \frac{\lbsset}{\cardset}$ be network-dependent parameters, then
\begin{align*}
    \!\! \com_3  = \graddim  \bigg( \! \nclients \! + \! \nclusters \! + \! \beta^\prime \sum_{\clientindex=1 }^\nclients \frac{\cardset + \maxcolluders}{\cardset}\bigg) < \bigg(\! 4\tau \! + \! \frac{\nclusters-\bscolluders}{\nclients+1} \! \bigg) \mincommr. \! %
\end{align*}
\end{theorem}

\begin{proof}
The total communication cost is decomposed as $\com_3 = \uecom_3 + \bscom_3 + \bsrcom_3 + \rcom_3 + \rfcom_3$, where the former is characterized by the secret sharing of each client $\clientindex$ through the base stations in $\bsset$ and the transmission of the random key $\randvec$ of dimension $\graddim$:
\begin{equation}
    \uecom_3 = \sum_{\clientindex \in  [\nclients]} \frac{\graddim \vert \bsset \vert}{\vert \bsset \vert - \maxcolluders} + \graddim \nclients,
\end{equation}
which can be upper and lower-bounded considering the worst and best-case connectivities of each of the clients as
\begin{equation}
    \graddim \nclients \frac{\maxcard}{\maxcard-\maxcolluders} + \graddim \nclients \leq \uecom_3 \leq \graddim \nclients (\maxcolluders + 1) + \graddim \nclients.
\end{equation}
The aggregation of the random keys $\randvec$ requires the base station to sequentially transmit the partial sum of keys among each other. Hence,
\begin{equation}
    \bscom_3 = \graddim (\nclusters-1). \label{eq:bsbs2}
\end{equation}
If multiple clients $\clientindex \in \commonbs[\tmpbs]$ share the same set $\bsset = \tmpbs$ with aligned evaluation points, the base stations $\clusterindex \in \tmpbs$ can aggregate the shares of those clients. Additionally considering the aggregated keys that must be transmitted, the communication cost from base stations to relay nodes reads as
\begin{align*}
    \bsrcom_3 = \graddim + \sum_{\tmpbs\in 2^{[\maxcard]} : \vert \tmpbs \vert > \maxcolluders} \mathds{1}\{\exists \clientindex: \bsset = \tmpbs\} \cdot \frac{d \vert \tmpbs \vert}{\vert \tmpbs \vert - \maxcolluders}.
\end{align*}
This quantity is minimized if all clients share the same set of base stations $\bsset$ and maximized if the $\bsset$'s are different for all clients $\clientindex \in [\nclients]$, i.e.,
\begin{align*}
    \bsrcom_3 &\geq \graddim + \extstart \graddim \extend \max_{\clientindex \in [\nclients]} \frac{\vert \bsset \vert}{\vert \bsset \vert - \maxcolluders} \geq \graddim + \graddim \frac{\maxcard}{\maxcard - \maxcolluders} \text{, and } \\
    \bsrcom_3 &\leq \graddim + \graddim \sum_{\clientindex = 1}^\nclients \frac{\vert \bsset \vert}{\vert \bsset \vert - \maxcolluders} \leq \graddim + \graddim \nclients (\maxcolluders + 1).
\end{align*}
The relay nodes do not employ their interconnection in the given setting, hence $\rcom_3 = 0$. The amount of communication from relay nodes to the federator is characterized by the aggregation opportunities for the base stations and the relay nodes. Since the relay nodes can simply forward the received shares and all clients with equal $\bsset$ also share $\relayset$, the following bound trivially holds $\rfcom_3 \leq \bsrcom_3$.
More specifically, we get
\begin{align*}
    \rfcom_3 &= \graddim + \sum_{\tmprelay\in 2^{[\maxcard]} : \vert \tmprelay \vert > \maxcolluders} \mathds{1}\{\exists \clientindex: \relayset = \tmprelay\} \cdot \frac{d \vert \tmprelay \vert}{\vert \tmprelay \vert - \maxcolluders} \\
    &= \graddim + \graddim \sum_{\clientindex = 1}^\nclients \frac{1}{n_{\vert \relayset \vert}} \frac{\vert \relayset \vert}{\vert \relayset \vert - \maxcolluders} 
    \leq \graddim + \graddim \sum_{\clientindex = 1}^\nclients \frac{\vert \relayset \vert}{\vert \relayset \vert - \maxcolluders}, \\ %
    \rfcom_3 &\geq \graddim + \frac{\graddim}{\nclients} \sum_{\clientindex = 1}^\nclients \frac{\vert \relayset \vert}{\vert \relayset \vert - \maxcolluders}, %
\end{align*}
where $\nclients_{\vert \relayset \vert}$ is the number of clients sharing $\relayset$ and the latter bound is by assuming that $\nclients_{\vert \relayset \vert} \leq \nclients$.
Oblivious to the actual connectivity pattern and assuming that there exists only one set of relays for all base station sets with equal cardinality, we obtain the following lower bound:
\begin{align*}
    \rfcom_3 &= \graddim + \graddim \sum_{r = \maxcolluders + 1}^{\maxcard} \mathds{1}\{\exists \clientindex: \vert \relayset \vert = r\} \cdot \frac{r}{r - \maxcolluders} \\
    &\geq \graddim + \graddim \frac{\maxcard}{\maxcard-\maxcolluders}.
\end{align*}

With $H_n$ being the harmonic number, a corresponding upper bound accounts to
\begin{align*}
    \rfcom_3 &= \graddim + \graddim \sum_{r = \maxcolluders + 1}^{\maxcard} \mathds{1}\{\exists \clientindex: \vert \relayset \vert = r\} \cdot \frac{r}{r - \maxcolluders} \\
    &\leq \graddim + \graddim \sum_{r = \maxcolluders + 1}^{\maxcard} \frac{r}{r - \maxcolluders}
    = \graddim + \graddim \sum_{r = 1}^{\extstart \maxcard - \maxcolluders \extend} \extstart \left(1 + \frac{\maxcolluders}{r}\right) \extend \\
    &= \graddim + \graddim (\maxcard - \maxcolluders - 1 + \maxcolluders H_{\extstart \maxcard - \maxcolluders \extend}) \\
    &= \graddim (\maxcard + \maxcolluders (H_{\extstart \maxcard - \maxcolluders \extend} - 1))
\end{align*}
To relate the achievability to the lower bound, we observe that $\nclients + \beta^\prime \sum_{\clientindex=1 }^\nclients \frac{\cardset+\maxcolluders}{\cardset} < 4 \sum_{\clientindex=1 }^\nclients \frac{\maxcolluders+\cardset}{\cardset}$ and $\frac{(\cardset + \maxcolluders) \lbsset}{(\lbsset + \bscolluders) \cardset} \leq \frac{\lbsset}{\cardset} \leq \tau$. Bounding $\mincommr$ from below by $ \frac{(\nclients + 1) \extstart \graddim \extend \nclusters}{\nclusters - \bscolluders}$, yields $\frac{\extstart \graddim \extend \nclusters}{\mincommr} = \frac{\nclusters- \bscolluders}{\nclients + 1}$. 
We conclude by merging all results and using $\nclusters \geq \maxcard$.
\end{proof}

We assumed the shares are simply forwarded from base stations to relay nodes. However, whether they can be forwarded without further action depends on the privacy requirements. If $\bscolluders = \relaycollusions$, base station $\clusterindex \in \commonbs[\tmpbs]$ can indeed directly forward the share of client $\clientindex \in \tmpbs$ to relay node $\relayidx_{\clientindex, \clusterindex}$. If $\relaycollusions < \bscolluders$, the shares forwarded to the relay nodes can have a lower level of privacy reflected by a smaller collusion resistance, which implies a potentially smaller share size or allows for a smaller set $\relayset$ that can be achieved by a transformation of the secret sharing scheme. Such a transformation requires a significant communication effort, which is beneficial if $\vert \relayset \vert < \vert \bsset \vert$ and direct forwarding is impossible. If $\relaycollusions > \bscolluders$, it is required to increase the level of privacy by transforming the secret sharing. Further, a transformation of secret shares can help align evaluation points to facilitate aggregations and thereby reduce the communication cost on higher levels of the hierarchy. We elaborate on such transformations in the sequel.

\section{Transforming Secret Sharings} \label{sec:transform}

This section investigates how to transform a McEliece-Sarwate secret sharing scheme \cite{mceliece1981sharing}  with parameters $(\tssn, \tssk, \tssz)$ into a different $(\tssn^\prime, \tssk^\prime, \tssz^\prime)$ secret sharing. The methods can be applied to cope with different circumstances on different layers of the hierarchy, e.g., the base stations and the relay nodes, but are likewise of independent interest. Similar ideas have appeared for different applications, c.f. \cite{goyal2022sharing} and references therein. In the following, we explicitly construct methods for hierarchical and wireless federated learning in the network topologies under consideration.

\subsection{General Transformation}
The following elaborations are not required if $\tssk=\tssk^\prime$ and $\tssz=\tssz^\prime$. Details for this case can be found in \cref{subsec:change_n_shares}.
For simplicity, we consider a single set of $\nbs$ base stations, all connected to a set of $\nr$ relays, which are fully connected. %
At the start, a secret message $\fullpaddedgrads \in \mathbb{F}^{\graddim}$ is shared among the base stations using an $(\nbs, \kbs, \bscolluders)$ threshold secret sharing scheme and the goal is to have the same message shared among the relay nodes with an $(\nr, \kr, \relaycollusions)$ threshold secret sharing scheme.%

In the setting considered in the rest of this paper, the secret message $\fullpaddedgrads$ represents an aggregate of gradient vectors padded with keys $\model + \randvec$ or an aggregate of keys $\randvec$.
The secret message is split into $\bssplits \defeq \kbs - \bscolluders$ parts of size $\frac{\graddim}{\bssplits}$ for the base stations and into $\rsplits \defeq \kr - \relaycollusions$ parts of size $\frac{\graddim}{\rsplits}$ for the relay nodes, i.e. \begin{align*}
\fullpaddedgrads = (\bspaddedgrads[1] \dots \bspaddedgrads[\bssplits]) = ({\rpaddedgrads[1]} \dots \rpaddedgrads[\rsplits]).
\end{align*}
Let $\bssecrand[1], \dots, \bssecrand[\bscolluders]$ be vectors drawn independently and uniformly at random from $\mathbb{F}_q^{\frac{\graddim}{\bssplits}}$ and $\rsecrand[1], \dots, \rsecrand[\relaycollusions]$ from \extstart $\mathbb{F}_q^{\frac{\graddim}{\rsplits}}$. \extend
Each base station $\bsindex \in \mathcal{Y}$ 
has an evaluation of the polynomial\begin{align*}
  \fold(x) = \sum_{\tmpindex \in [\bssplits]} x^{\tmpindex-1} \bspaddedgrads[\tmpindex] + \sum_{\tmpindex \in [\bscolluders]} x^{\tmpindex + \bssplits - 1} \bssecrand[\tmpindex] \extstart \in \mathbb{F}_q^{\frac{\graddim}{\bssplits}} \extend
\end{align*}
at some unique non-zero evaluation point $\bseval_\bsindex$ and the goal is for each relay node $\relayidx$ to have an evaluation of a different polynomial\begin{align*}
  \fnew(x) = \sum_{\tmpindex \in [\rsplits]} x^{\tmpindex-1} \rpaddedgrads[\tmpindex] + \sum_{\tmpindex \in [\relaycollusions]} x^{\tmpindex + \rsplits - 1} \rsecrand[\tmpindex] \extstart \in \mathbb{F}_q^{\frac{\graddim}{\rsplits}} \extend
\end{align*}
at a unique non-zero evaluation point $\relayeval_\relayidx$.

The shares of any $\kbs$ base stations contain complete information about $\fullpaddedgrads$. Without loss of generality, we designate base stations $1, \dots, \kbs$ to carry out the secure aggregation. 
The shares $\fnew(\relayeval_\relayidx)$ among relay nodes can be expressed as a weighted sum of the  shares $\fold(\bseval_\bsindex)$ of base stations $1, \dots, \kbs$, and some independent random numbers.

  To formally state this relation, we introduce some notation. 
  Define the Vandermonde matrix
  $\vanderbstrunk \defeq (\extstart \bseval_\bsindex^{\tmpindex-1} \extend)_{\tmpindex \in [\kbs], \bsindex\in [\kbs]}$ 
  and let $\vbsinvtrunk \in \mathbb{F}^{\frac{\kbs\graddim}{\bssplits} \times \graddim}$ consist of the first $\graddim$ columns of \extstart $\vanderbstrunk^{-1} \otimes \eye[\frac{\graddim}{\bssplits}]$,\extend where $\eye[\zeta]$ is the $\zeta \times \zeta$ identity matrix.
  For all $\relayidx\in [\nr]$ let\begin{align*}
    \veval &\defeq (1\ \relayeval_\relayidx\ \relayeval_\relayidx^2\ \dots\ \relayeval_\relayidx^{\rsplits-1}), \\ 
    \vevalr &\defeq (\relayeval_\relayidx^{\rsplits}\ \relayeval_\relayidx^{\rsplits+1}\ \dots\ \relayeval_\relayidx^{\kr-1}), \\ 
    \vrb[\relayidx] &\defeq \extstart {\vevalr}^T \otimes \eye[\frac{\graddim}{\rsplits}], \extend \\ 
    \weightmat[\relayidx] &\defeq \vbsinvtrunk\cdot (\extstart \veval[\relayidx]^T \otimes \eye[\frac{\graddim}{\rsplits}] \extend) \in \mathbb{F}^{\frac{\kbs\graddim}{\bssplits} \times \frac{\graddim}{\rsplits}},
  \end{align*}
  and let $\weightmat[\relayidx, \bsindex] \in \mathbb{F}^{\frac{\graddim}{\bssplits} \times \frac{\graddim}{\rsplits}}$ be submatrices of $\weightmat[\relayidx]$, s.t. $\weightmat[\relayidx] = \left(\weightmatT[\relayidx, 1]\ \weightmatT[\relayidx, 2]\ \dots\ \weightmatT[\relayidx, \kbs]\right)^T$.

\begin{proposition}
  The share of each relay node $\relayidx \in [\nr]$ can be expressed as
  \begin{align*}
    \fnew(\relayeval_\relayidx) &= \sum_{\bsindex \in [\kbs]} \fold(\bseval_\bsindex)\weightmat[\relayidx, \bsindex]  + \extrarand[\bsindex]\vrb[\relayidx], 
  \end{align*}
  where the vectors $\extrarand[\bsindex] \in \mathbb{F}^{\frac{\graddim}{\rsplits}}, \bsindex \in [\kbs]$ are drawn independently and uniformly at random.
\end{proposition}
\begin{proof}
  Let $\fullbssecrand \defeq (\bssecrand[1] \dots\ \bssecrand[\bscolluders])$ and $\fullrsecrand \defeq (\rsecrand[1] \dots\ \rsecrand[\relaycollusions])$.
  We can express the vector of the $\kbs$ base stations' concatenated shares $\foldevalstrunk \defeq (\fold(\bseval_1) \dots\ \fold(\bseval_\kbs))$ as\begin{align*}
    \foldevalstrunk &= (\fullpaddedgrads\ \fullbssecrand) \cdot (\extstart \vanderbstrunk \otimes \eye[\frac{\graddim}{\bssplits}] \extend)%
  \end{align*}
  The Vandermonde matrix $\vanderbstrunk$ is invertible, hence the inverse of \extstart $\vanderbstrunk \otimes \eye[\frac{\graddim}{\bssplits}]$ is $\vanderbstrunk^{-1} \otimes \eye[\frac{\graddim}{\bssplits}]$, \extend and $\fullpaddedgrads = \foldevalstrunk\cdot \vbsinvtrunk$.
  Using this, we have 
  \begin{align*}
    \fnew(\relayeval_\relayidx) &= \sum_{\tmpindex \in [\rsplits]} \relayeval_\relayidx^{\tmpindex-1} \rpaddedgrads[\tmpindex] + \sum_{\tmpindex \in [\relaycollusions]} \relayeval_\relayidx^{\tmpindex + k^\prime - \relaycollusions - 1} \rsecrand[\tmpindex] \\
                                &= \sum_{\tmpindex \in [\rsplits]} \rpaddedgrads[\tmpindex]\eye[\frac{\graddim}{\rsplits}] \relayeval_\relayidx^{\tmpindex-1} + \sum_{\tmpindex \in [\relaycollusions]}  \rsecrand[\tmpindex]\eye[\frac{\graddim}{\rsplits}] \relayeval_\relayidx^{\tmpindex + k^\prime - \relaycollusions - 1}\\
                                &= \fullpaddedgrads\cdot \extstart(\veval^T \otimes \eye[\extstart \frac{\graddim}{\rsplits} \extend])\extend + \fullrsecrand\cdot \extstart({\vevalr}^T \otimes \eye[\frac{\graddim}{\rsplits}])\extend \\
                                &= \foldevalstrunk\cdot \vbsinvtrunk\cdot \extstart(\veval^T \otimes \eye[\frac{\graddim}{\rsplits}])\extend + \big(\sum_{\bsindex \in [\nbs]} \extrarand[\bsindex] \big) \cdot \vrb[\relayidx] \\
                                &= \sum_{\bsindex \in [\kbs]} \fold(\bseval_\bsindex)\weightmat[\relayidx, \bsindex] + \extrarand[\bsindex] \vrb[\relayidx],
  \end{align*}
  which concludes the proof.
\end{proof}

The share of each relay node \extstart $\tmprelayidxa \in [\nr]$ \extend is constructed separately through the following procedure. Each base station $\bsindex \in [\kbs]$ computes its contribution to relay node $\tmprelayidxa$'s share\begin{align*}
  \contribution[\tmprelayidxa,\bsindex] \defeq \fold(\bseval_\bsindex) \weightmat[\tmprelayidxa, \bsindex] + \extrarand[\bsindex] \vrb[\tmprelayidxa] \extstart \in \mathbb{F}^{\frac{\graddim}{\rsplits}} \extend.
\end{align*}
It splits its contribution vector $\contribution[\tmprelayidxa, \bsindex]$ into $\nr - \relaycollusions$ equally sized parts and constructs a \emph{transport polynomial} 
$\ftransport[\tmprelayidxa, \bsindex](x) \defeq \sum_{\partindex \in [\nr-\relaycollusions]} \contribution[\tmprelayidxa,\bsindex,\partindex] x^{\partindex-1} + \sum_{\tmpindex \in \relaycollusions} \tsecrand_{\bsindex,\tmpindex} x^{\tmpindex+\nr-\relaycollusions-1}$, where \extstart $\tsecrand_{\bsindex,\tmpindex} \in \mathbb{F}^{\frac{\graddim}{\rsplits (\nr-\relaycollusions)}}$ \extend are drawn independently and uniformly at random.
Base station $\bsindex$ sends one evaluation $\ftransport[\tmprelayidxa,\bsindex](\relayeval_\tmprelayidxb)$ to each relay node $\tmprelayidxb \in [\nr]$, which sums the shares from all base stations and sends $\sum_{\bsindex \in [\kbs]} \ftransport[\tmprelayidxa,\bsindex](\relayeval_\tmprelayidxb)$ to relay node $\tmprelayidxa$.
Having received $\nr$ evaluations of $\ftransport[\tmprelayidxa](x) \defeq \sum_{\bsindex \in [\kbs]} \ftransport[\tmprelayidxa,\bsindex](x)$, relay node $\tmprelayidxa$ can interpolate the polynomial and decode its share. 
The communication cost is $$d \frac{\nr^2 (\kbs+1) - \nr}{\extstart(\kbs-\bscolluders)\extend(\nr-\relaycollusions)}.$$

\subsection{Changing the Number of Shares} \label{subsec:change_n_shares}
In the simplest case, the robustness against collusions and the threshold of the secret sharing schemes among base stations and relays are the same, and the number of relays is less than or equal to the number of base stations, i.e., $\bscolluders=\relaycollusions$, $\kbs=\kr$ and $\nbs\geq \nr$. In this case, only the first $\nr$ base stations forward their shares to distinct relays. 
The communication cost of this approach is $\nr \frac{d}{\rsplits}$ symbols in $\mathbb{F}_q$.

If $n < n^\prime$, each base station must forward their share to a distinct relay node and $n^\prime - n$ additional shares can be computed as in the general case. Hence, the communication cost can be calculated to \extstart $$d \frac{\nbs}{\kbs - \bscolluders} + d \frac{(\nr - \nbs) (\kbs \nr + (\nr -1))}{(\kbs - \bscolluders) (\nr - \relaycollusions)}.$$ \extend

Such concepts can, for example, be used to adapt the dropout resilience to the specific requirements of each layer.

\begin{remark}[Communication Efficient Secret Sharing]
  In the setting of interest, $n$ is expected to be close to $k$. However, in general, when transforming one secret sharing into another one and if $n$ is significantly larger than $k$, communication efficient secret sharing~\cite{huang2016communication,bitar2018staircase} can be employed to further reduce the overall communication.
\end{remark}

\section{Conclusion} \label{sec:conclusion}
In this work, we introduced the problem of private aggregation for wireless and hierarchical federated learning scenarios. We proposed information-theoretic private aggregation schemes under full and partial collusion assumptions for the clients, base stations, relay nodes, and the federator. We analyzed the communication cost depending on the network topology and the collusion assumptions. We derived fundamental limits in terms of communication cost and showed that our proposed schemes are at most a multiplicative factor away. Since requirements w.r.t. privacy and dropout resilience might differ on different levels in the hierarchy, we investigated the transformation of secret sharing schemes. %

The communication cost of such aggregation schemes could further be reduced by loosening the privacy guarantee to non-zero leakage of the clients' data. Additionally, the communication cost and its importance might differ across the hierarchy levels. Designing schemes that respect such objectives is left for future work.  

\clearpage 

\appendix
\extstart
\subsection{Proof of~\cref{prop:sharesize}} \label{app:proof_of_sharesize}
\begin{proof}
  Let $\tssind_1, \dots, \tssind_{\tssk} \in [\tssn]$ be distinct and all logarithms be to the base $q$. 
  Assume $\tsssecret$ is uniformly distributed over $\mathbb{F}_q^{\tssdim}$, i.e., $\entropy{\tsssecret} = \tssdim$.
  We first prove $\entropy{\tssshare_{\tssind_1}, \dots, \tssshare_{\tssind_{\tssk-\tssz}}} \geq \entropy{\tsssecret}$, from which the result will follow. We have
  \begin{align*}
    \entropy{\tssshare_{\tssind_1}, \dots, \tssshare_{\tssind_{\tssk-\tssz}}}
      &\geq \entropy{\tssshare_{\tssind_1}, \dots, \tssshare_{\tssind_{\tssk-\tssz}} \given \tssshare_{\tssind_{\tssk-\tssz+1}}, \dots, \tssshare_{\tssind_{\tssk}}} \\ 
      &\geq \mathrm{I}(\tsssecret; \tssshare_{\tssind_1}, \dots, \tssshare_{\tssind_{\tssk-\tssz}} \given \tssshare_{\tssind_{\tssk-\tssz+1}}, \dots, \tssshare_{\tssind_{\tssk}}) \\
      &= \entropy{\tsssecret \mid \tssshare_{\tssind_1},\dots, \tssshare_{\tssind_z}} - \entropy{M \mid \tssshare_{\tssind_1},\dots, \tssshare_{\tssind_\tssk}} \\
      &= \entropy{\tsssecret}.
  \end{align*}
  Let $\tssinds \defeq \{\tsstmpset \subset [\tssn] : \vert tsstmpset\vert  = \tssk-\tssz\}$ consist of all subsets of cardinality $\tssk-\tssz$ of $[\tssn]$. We have
  \begin{align*}
    {\tssn \choose {\tssk-\tssz}} \entropy{\tsssecret} &\leq
    \sum_{\{\tssind_1, \dots, \tssind_{\tssk-\tssz}\} \in \tssinds} \entropy{\tssshare_{\tssind_1}, \dots, \tssshare_{\tssind_{\tssk-\tssz}}} \\
   &\leq \sum_{\{\tssind_1, \dots, \tssind_{\tssk-\tssz}\} \in \tssinds} \entropy{\tssshare_{\tssind_1}} + \dots + \entropy{\tssshare_{\tssind_{\tssk-\tssz}}} \\
      &\stackrel{(a)}{=} \sum_{\tssind \in [\tssn]} {{\tssn-1} \choose {\tssk-\tssz-1}}  \entropy{\tssshare_{\tssind}} \\
      &\leq {{\tssn-1} \choose {\tssk-\tssz-1}} \sum_{\tssind \in [\tssn]}  \tsssharedim_\tssind, 
  \end{align*}
  where $(a)$ holds since every index $\tssind \in [\tssn]$ occurs in ${\tssn -1} \choose {\tssk-\tssz-1}$ elements of $\tssinds$.
    
  And finally,
  \begin{align*}
    \sum_{\tssind \in [\tssn]}  \tsssharedim_\tssind \geq \tssdim \frac{{\tssn \choose {\tssk-\tssz}}}{{{\tssn-1} \choose {\tssk-\tssz-1}}} = \tssdim \frac{\tssn}{\tssk-\tssz}.
  \end{align*}
\end{proof}
\extend

\subsection{Proof of~\cref{thm:IT-private_scheme}} \label{app:proof_of_thm_IT-private_scheme}

\begin{proof}%

For convenience, we define the following sets:
\begin{align*}
  \Submodels[\tmpset] &\defeq \{\submodel[j]: \clientindex \in \tmpset, j \in [\bsaddconn]\}, \\
\mathcal{F}^{\tmpset} &\defeq \{\clientpoly[\alpha_\clusterindex]: \clientindex \in [\nclients], \clusterindex \in \tmpset \subseteq [\nclusters], \given \tmpset \given \leq \bscolluders\}, \\
\mathcal{K}^\mathcal{\tmpset} &\defeq \{\extstart \randvec^{(j)} \extend: \clientindex \in \tmpset \extstart\subseteq [n]\extend, j \in [\bsaddconn]\}, \\
\Sumvecs &\defeq \{\submodel + \extstart \randvec^{(j)} \extend: \clientindex \in [\nclients], j \in [\bsaddconn]\}.
\end{align*}

\extstart
With these definitions we have $\observations^{\uecolluderset} = \Submodels[\uecolluderset] \cup \Randvecs[\uecolluderset] \cup \aggregatedgradient$ and $\observations^{\mathcal{B}} = \Obs$.
\extend

We start by proving that the constructed scheme satisfies the privacy constraint stated in \eqref{eq:it_privacy} as follows.

\extstart
\begin{align}
    \mathrm{I}&\left(\observations^{\mathcal{C}}, \observations^{\mathcal{B}}; \privatedata^{\noncuriousset} \mid \aggregatedgradient, \privatedata^{\uecolluderset} \right) \nonumber \\
    &= I\left(\Randvecscol, \Submodels[\mathcal{C}], \Obs, \aggregatedgradient; \privatedata^{\noncuriousset} \mid \aggregatedgradient, \privatedata^{\uecolluderset} \right) \nonumber \\
    &= I\left(\Randvecscol, \Obs; \privatedata^{\noncuriousset} \mid \aggregatedgradient, \privatedata^{\uecolluderset} \right) \nonumber \\
    &\leq I\left(\Randvecs, \Obs; \privatedata^{\noncuriousset} \mid \aggregatedgradient, \privatedata^{\uecolluderset} \right) \nonumber \\
    &= I\left(\Obs; \privatedata^{\noncuriousset} \mid \Randvecs, \aggregatedgradient, \privatedata^{\uecolluderset} \right) + I\left(\Randvecs; \privatedata^{\noncuriousset} \mid \aggregatedgradient, \privatedata^{\uecolluderset} \right) \nonumber \\
    &\stackrel{(a)}{=}  I\left(\Obs; \privatedata^{\noncuriousset} \mid \Randvecs, \aggregatedgradient, \privatedata^{\uecolluderset} \right) \nonumber \\
    &= I\left(\Obs; \privatedata^{\noncuriousset}, \Randvecs, \aggregatedgradient, \privatedata^{\uecolluderset} \right) - I\left(\Obs; \Randvecs, \aggregatedgradient, \privatedata^{\uecolluderset} \right) \nonumber \\ \nonumber
    &\leq I\left(\Obs; \privatedata^{\uecolluderset}, \privatedata^{\noncuriousset}, \Randvecs, \aggregatedgradient \right) \\ \nonumber
    &= I\left(\Obs; \privatedata^{\clientset}, \Randvecs \right) \\ \nonumber
    &\leq I\left(\Obs; \privatedata^{\clientset}, \Randvecs, \Sumvecs \right) \\ \nonumber
    &\stackrel{(b)}{=} I\left(\Obs; \Sumvecs \right) \\ \nonumber
    &= 0, \nonumber
\end{align}
where (a) holds because the random vectors in \Randvecs are independent of all gradients and (b) holds because $(\Submodels, \Randvecs) \rightarrow \Sumvecs \rightarrow \Obs$ form a Markov chain.
\extend

Next, we prove privacy against at most $\uecolluders$ clients colluding with the federator. For the proposed scheme, we have %
\begin{align}
    \mathrm{I}&\left(\observations^F, \observations^{\mathcal{C}}; \privatedata^{\noncuriousset} \mid \aggregatedgradient, \privatedata^{\uecolluderset} \right) \nonumber \\
    &\stackrel{(c)}{\leq} \I{\Sumvecs, \aggregatedgradient, \aggregatedkeys, \Submodels[\uecolluderset], \Randvecscol; \Submodels[\noncuriousset] \mid \aggregatedgradient, \Submodels[\uecolluderset]} \nonumber \\
    &=\I{\Sumvecs, \aggregatedkeys, \Randvecscol; \Submodels[\noncuriousset] \mid \aggregatedgradient, \Submodels[\uecolluderset]} \nonumber \\
    &= \entropy{\Sumvecs, \aggregatedkeys, \Randvecscol \mid \aggregatedgradient, \Submodels[\uecolluderset]} - \entropy{\Sumvecs, \aggregatedkeys, \Randvecscol \mid \aggregatedgradient, \Submodels[\uecolluderset], \Submodels[\noncuriousset]} \nonumber \\
    &\stackrel{(d)}{=} \entropy{\Sumvecs \mid \aggregatedgradient, \Submodels[\uecolluderset]} - \entropy{\Sumvecs \mid \aggregatedgradient, \Submodels[\uecolluderset], \Submodels[\noncuriousset]} \nonumber \\
    & \stackrel{(e)}{=} \entropy{\Sumvecs} - \entropy{\Sumvecs} \label{eq:privacy_padded_gradients}
    = 0, %
\end{align}
where the inequality in (c) follows since the set of random variables in $\observations^F, \observations^{\mathcal{C}}$ is a subset of the random variables in $\Sumvecs, \aggregatedgradient, \aggregatedkeys, \Submodels[\uecolluderset], \Randvecscol$. The equality in $(d)$ follows because $\aggregatedkeys$ and $\Randvecscol$ are deterministic functions of $\Sumvecs$ and $\aggregatedgradient, \Submodels[\uecolluderset]$. The equality in (e) holds because each element in $\Sumvecs$ is the sum of a $\model$ with a distinct random vector whose entries are drawn independently and uniformly at random from $\mathbb{F}_q$. By Shannon's one-time pad, the entries in $\Sumvecs$ are independent of all the $\model$'s. Using the non-negativity of mutual information concludes the proof of \cref{thm:IT-private_scheme}.
\end{proof}

\subsection{Proof of \cref{thm:IT-private_scheme_communication}} \label{app:proof_of_thm_IT-private_scheme_communication}

\begin{proof}[Proof of \cref{thm:IT-private_scheme_communication}]
We next analyze the communication cost of the introduced scheme, which depends on the connectivity pattern $\bsclient$ of each client. The total communication cost $\com_1$ is composed of the communication from clients to the base stations $\uecom_1$, the communication between base stations $\bscom_1$, and the communication from the base stations to the federator $\bsfcom_1$, i.e., $\com_1 = \uecom_1 + \bscom_1 + \bsfcom_1$. 

Each client $\clientindex$ needs to send a secret sharing of their computed gradient to all base stations in $\bsclient$ which incurs a communication cost of $\graddim\frac{\vert \bsclient \vert}{\vert \bsclient \vert - \bscolluders}$ symbols. In addition, each client $\clientindex$ sends the random vector $\randvec$ of length $\graddim$ to any base station. Hence, $\uecom_1$ can be expressed as
\begin{equation}
    \uecom_1 = \sum_{\clientindex \in  [\nclients]} \frac{\graddim \vert \bsclient \vert}{\vert \bsclient \vert - \bscolluders} + \graddim \nclients. \label{eq:uebs}
\end{equation}
This objective is minimized for large $\vert\bsclient\vert$'s. A lower bound can be found when all clients are connected to all base stations, i.e., $\forall \clientindex \in [\nclients]$ we have that $\vert \bsclient \vert = \nclusters$. An upper bound results from the scenario where each client is only connected to $\bscolluders+1$ base stations. Formally,
\begin{equation*}
    \graddim \nclients \frac{\nclusters}{\nclusters-\bscolluders} + \graddim \nclients \leq \uecom_1 \leq \nclients\graddim (\bscolluders+1) + \graddim \nclients.
\end{equation*}

To compute and share the aggregated random vectors $\aggregatedkeys$ with the federator, the base stations send their local aggregated random vectors sequentially to the next base station. The last base station shares the result with the federator. Hence, 
\begin{equation}
    \bscom_1 = \graddim (\nclusters-1). \label{eq:bsbs}
\end{equation}

The base stations need to forward the secret shares to the federator. Recall that if a set of clients $\commonconn$ share the same connectivity pattern $\tmpconn$, then base stations in $\tmpconn$ simply send the aggregate (sum) of the shares of those clients to the federator. The total communication cost from the base stations to the federator is the communication incurred by each connectivity pattern given by $\graddim\frac{\vert \tmpconn \vert}{\vert \tmpconn \vert - \bscolluders}$. In addition, one base station must send the aggregated random vector $\aggregatedkeys$ to the federator. Hence, the communication cost from base stations to the federator is expressed as 
\begin{equation}
    \bsfcom_1 = \graddim  + \!\!\!\!\!\!\!\!\! \sum_{\tmpconn \in 2^{[\nclusters]}: \vert \tmpconn \vert > \bscolluders} \!\!\!\!\!\!\!\!\!\! \mathds{1}\{\exists \clientindex: \bsclient = \tmpconn \} \cdot \frac{\graddim \vert \tmpconn \vert}{\vert \tmpconn \vert - \bscolluders}, \label{eq:bsf}
\end{equation}
where $2^{[\nclusters]}$ is the power set of $[\nclusters]$ giving all possible connectivity patterns.
In the best case, all clients share the exact same connectivity pattern, i.e., for all clients $\clientindex_1,\clientindex_2$ it holds that $\bsclientone = \bsclienttwo$. In this case, the communication cost from the base stations to the federator is minimized. Each base station has to send exactly one vector of size $\frac{\graddim}{\vert\bsclient\vert - \bscolluders}$, the sum of all received evaluations, to the federator. This is due to the properties of secret sharing, so the total communication cost is $\frac{\graddim \vert\bsclient\vert}{\vert\bsclient\vert - \bscolluders}$. 
The communication cost is further decreased if all clients are connected to all base stations, i.e., $\vert \bsclient \vert = \nclusters$. %
In the worst case, however, the connectivity patterns of all clients are pairwise distinct, i.e., $\forall \clientindex_1,\clientindex_2 \in [\nclients], \clientindex_1 \neq \clientindex_2$ it holds that $\bsclient[\clientindex_1] \neq \bsclienttwo$. The communication is further increased if for all clients $\vert\bsclient\vert = \bscolluders+1$. In this case, for each connectivity pattern $\bsclient$ the base stations have to collectively send a vector of size $\frac{\graddim \vert \bsclient\vert}{\vert\bsclient\vert - \bscolluders} = \graddim (\bscolluders+1)$.
Hence, we obtain the following:
\begin{equation*}
    \graddim + \graddim \frac{\nclusters}{\nclusters-\bscolluders} \leq \bsfcom_1 \leq \graddim + \nclients \graddim (\bscolluders +1)
\end{equation*}

Using the bounds derived on $\uecom_1$ and $\bsfcom_1$, we can bound $\com_1$ as in \eqref{eq:loose_bound}.

We need the following derivations to obtain the tighter bound of \eqref{eq:tight_bound}. From \eqref{eq:bsf} it follows that
\begin{align}
    \!\bsfcom_1 & \geq \graddim + \max_{\clientindex \in [\nclients]} \frac{\graddim\vert \bsclient \vert}{\vert \bsclient \vert-\bscolluders} \geq \graddim + \frac{\graddim}{\nclients} \sum_{\clientindex \in [\nclients]} \frac{\vert \bsclient \vert}{\vert \bsclient \vert-\bscolluders}, \!\!\label{eq:intermediate}
\end{align}
since the communication cost of a connectivity pattern is a decreasing function of $\vert\bsclient\vert$ and where the second inequality follows since the maximum of $\frac{\vert \bsclient \vert}{\vert \bsclient \vert-\bscolluders}$ for $\clientindex \in [\nclients]$ is larger than its average.
From \eqref{eq:bsf}, observing that $\bsfcom_1$ is maximized if all the connectivity patterns of the clients are distinct yields the following upper bound on $\bsfcom_1$:
\begin{align}
    \bsfcom_1 & \leq \graddim + \graddim \sum_{\clientindex \in [\nclients]} \frac{\vert \bsclient \vert}{\vert \bsclient \vert - \bscolluders} \label{eq:intermediate_upper_bound}
\end{align}

Combining \eqref{eq:uebs}, \eqref{eq:bsbs}, \eqref{eq:intermediate} and \eqref{eq:intermediate_upper_bound} and recalling that $\com_1 = \uecom_1+\bscom_1+\bsfcom_1$ we obtain the following bound on $\frac{1}{\graddim} (\com_1 -\graddim \nclients-\graddim \nclusters)$:
\begin{align*}
    \frac{\nclients+1}{\nclients} \sum_{\clientindex \in [\nclients]} \frac{\vert \bsclient \vert}{\vert \bsclient \vert-\bscolluders} \leq \frac{\com_1 -\graddim \nclients-\graddim \nclusters}{\graddim} \leq 2\cdot \sum_{\clientindex \in [\nclients]} \frac{\vert \bsclient \vert}{\vert \bsclient \vert - \bscolluders}
\end{align*}

Consequently, $\com_1 -\graddim \nclients-\graddim \nclusters$ is a $\beta$-fraction of $\graddim \cdot \sum_{\clientindex \in [\nclients]} \frac{\vert \bsclient \vert}{\vert \bsclient \vert - \bscolluders}$, where $\frac{\nclients+1}{\nclients} \leq \beta \leq 2$. The factor $\beta$ depends on the relation between the connectivity parameters $\bsclient$ for all clients $\clientindex \in [\nclients]$. %

Lastly, we prove the relation to the lower bound $\mincomm$. Noting that $\frac{\bscolluders + \bsaddconn}{\bsaddconn}\geq \frac{\nclusters}{\nclusters-\bscolluders}$ for all $\clientindex\in [\nclients]$, from \cref{thm:converse} we obtain $\mincomm \geq (\nclients + 1) \frac{\extstart \graddim \extend \nclusters}{\nclusters - \bscolluders}$, which yields $\frac{\extstart \graddim \extend \nclusters}{\mincomm} \leq \frac{\nclusters- \bscolluders}{\nclients + 1}$. Observing that $\nclients + \beta \sum_{\clientindex=1 }^\nclients \frac{\bscolluders+\bsaddconn}{\bsaddconn} < 3 \sum_{\clientindex=1 }^\nclients \frac{\bscolluders+\bsaddconn}{\bsaddconn}$ and combining both bounds, we conclude the proof.
\end{proof}

\subsection{Proof of~\cref{thm:IT-private_scheme_extended}} \label{app:proof_of_thm_IT-private_scheme_extended}

\begin{proof}
To prove the privacy guarantee, we require the definitions in the proof of \cref{thm:IT-private_scheme}. Additionally, we introduce
\begin{align*}
    \mathcal{D}^{\tmpset} \defeq \Big\{\sum_{\clientindex \in \mathcal{I}} \clientpoly[\alpha_\relayidx]: \relayidx \in \tmpset \subseteq [\ntotalrelays], \vert \tmpset \vert \leq \relaycollusions, \mathcal{I} \subseteq [\nclients]\Big\}    
\end{align*}
as the potential set of partially aggregated secret shares observed by a colluding set of at most $\relaycollusions$ relays.
The first condition $\mathrm{I}\left(\observations^{\mathcal{C}}, \observations^{\mathcal{B}}; \privatedata^{\noncuriousset} \vert \extstart\aggregatedgradient,\extend \privatedata^{\uecolluderset} \right) = 0$ is proven the same way as in \cref{thm:IT-private_scheme} and is omitted. We, therefore, continue to prove the second privacy condition
\begin{align}
    \mathrm{I}&\left(\observations^{\mathcal{C}}, \observations^{\mathcal{R}}, \observations^F; \privatedata^{\noncuriousset} \mid \aggregatedgradient, \privatedata^{\uecolluderset} \right) \nonumber \\
    &\stackrel{(a)}{\leq} \I{\Sumvecs, \mathcal{D}^{\mathcal{R}}, \aggregatedgradient, \aggregatedkeys, \Submodels[\uecolluderset], \Randvecscol; \Submodels[\noncuriousset] \mid \aggregatedgradient, \Submodels[\uecolluderset]} \nonumber \\
    &=\I{\Sumvecs, \mathcal{D}^{\mathcal{R}}, \aggregatedkeys, \Randvecscol; \Submodels[\noncuriousset] \mid \aggregatedgradient, \Submodels[\uecolluderset]} \nonumber \\
    &= \entropy{\Sumvecs, \mathcal{D}^{\mathcal{R}}, \aggregatedkeys, \Randvecscol \mid \aggregatedgradient, \Submodels[\uecolluderset]} - \entropy{\Sumvecs, \mathcal{D}^{\mathcal{R}}, \aggregatedkeys, \Randvecscol \mid \aggregatedgradient, \Submodels[\uecolluderset], \Submodels[\noncuriousset]} \nonumber \\
    &\stackrel{(b)}{=} \entropy{\Sumvecs, \mathcal{D}^{\mathcal{R}} \mid \aggregatedgradient, \Submodels[\uecolluderset]} - \entropy{\Sumvecs, \mathcal{D}^{\mathcal{R}} \mid \aggregatedgradient, \Submodels[\uecolluderset], \Submodels[\noncuriousset]} \nonumber \\
    &= \I{\Sumvecs, \mathcal{D}^{\mathcal{R}}; \Submodels[\noncuriousset] \mid \aggregatedgradient, \Submodels[\uecolluderset]} \nonumber \\
    &= \I{\Sumvecs; \Submodels[\noncuriousset] \mid \aggregatedgradient, \Submodels[\uecolluderset]} + \I{\mathcal{D}^{\mathcal{R}}; \Submodels[\noncuriousset] \mid \aggregatedgradient, \Submodels[\uecolluderset], \Sumvecs} \nonumber \\
    & \stackrel{(c)}{=} \entropy{\mathcal{D}^{\mathcal{R}} \mid \aggregatedgradient, \Submodels[\uecolluderset], \Sumvecs} - \entropy{\mathcal{D}^{\mathcal{R}} \mid \aggregatedgradient, \Submodels[\uecolluderset], \Submodels[\noncuriousset], \Sumvecs} \nonumber \\
    & \stackrel{(d)}{=} \entropy{\mathcal{D}^{\mathcal{R}} \mid \Sumvecs} - \entropy{\mathcal{D}^{\mathcal{R}} \mid \Sumvecs} = 0, \nonumber
\end{align}
where (a) holds since the observations are a subset of the random variables stated, (b) holds since the keys $\aggregatedkeys$ and $\Randvecscol$ are a deterministic function of $\aggregatedgradient$, $\Submodels[\uecolluderset]$ and $\Sumvecs$. We obtain (c) by using that $\I{\Sumvecs; \Submodels[\noncuriousset] \mid \aggregatedgradient, \Submodels[\uecolluderset]} = 0$, which we proved in \eqref{eq:privacy_padded_gradients}. (d) is due to data processing inequality, i.e., \extstart $\mathcal{D}^{\mathcal{R}}$ depends on $(\aggregatedgradient, \Submodels[\uecolluderset], \Submodels[\noncuriousset])$ through $\Sumvecs$ \extend, which is why removing the conditioning on any of these variables does not change the entropy $\entropy{\mathcal{D}^{\mathcal{R}} \mid \aggregatedgradient, \Submodels[\uecolluderset], \Submodels[\noncuriousset], \Sumvecs}$.
\end{proof}

\subsection{Proof of~\cref{thm:IT-private_scheme_seckey}} \label{app:proof_of_thm_IT-private_scheme_seckey}

\begin{proof}
To prove the privacy of the proposed extension, we require the following definitions in addition to the notation introduced in the proof of \cref{thm:IT-private_scheme}.
\begin{align*}
\gradpolyset^{\tmpset} &\defeq \{\clientpolygrads[\alpha_\clusterindex]: \clientindex \in [\nclients], \clusterindex \in \tmpset \subseteq [\nclusters], \vert \tmpset \vert \leq \bscolluders\}, \\
\keypolyset^{\tmpset} &\defeq \{\clientpolykeys[\alpha_\clusterindex^\prime]: \clientindex \in [\nclients], \clusterindex \in \tmpset \subseteq [\nclusters], \vert \tmpset \vert \leq \bscolluders\}, \\
\setpaddedsums &\defeq \Big\{\! \sum_{\clientindex \in \commonmain} \model + \randvec: \tmpmain \in \mainsets \Big\}, \\
\setgradientsums &\defeq \Big\{\! \sum_{\clientindex \in \commonsec} \randvec: \tmpsec \in \secsets\Big\}.
\end{align*}
Informally, $\gradpolyset^\tmpset$ denotes a set of at most $\bscolluders$ secret shares of padded gradients observed by base stations $\clusterindex \in \tmpset$ for each gradient, and $\keypolyset^\tmpset$ a set of at most $\bscolluders$ secret shares of random keys observed by base stations $\clusterindex \in \tmpset$ for each gradient. $\setpaddedsums$ denotes the set of all partial aggregations of padded gradients that the federator observes based on the clients' choices of $\bsmain$ and $\bssec$, and $\setgradientsums$ refers to all partially aggregated keys correspondingly. We can now continue to prove the theorem statement.

\begin{align*}
    &\mathrm{I}\left(\observations^F, \observations^\mathcal{B}, \observations^{\mathcal{C}}; \privatedata^{\noncuriousset} \vert \aggregatedgradient, \privatedata^{\uecolluderset} \right) \\
    &\leq \mathrm{I}\left(\gradpolyset^\mathcal{B}, \setpaddedsums, \keypolyset^\mathcal{B},  \setgradientsums, \aggregatedkeys, \Randvecscol; \privatedata^{\noncuriousset} \given \aggregatedgradient, \privatedata^{\uecolluderset} \right) \\
    &\!\overset{(a)}{=} \! \mathrm{I}\left(\gradpolyset^\mathcal{B}, \setpaddedsums, \keypolyset^\mathcal{B},  \setgradientsums, \Randvecscol; \privatedata^{\noncuriousset} \given \aggregatedgradient, \privatedata^{\uecolluderset} \right) \\
    &=  \I{ \gradpolyset^\mathcal{B}, \keypolyset^\mathcal{B},  \setgradientsums, \Randvecscol;\Submodels[\noncuriousset] \given \aggregatedgradient, \Submodels[\uecolluderset]} \!+\! \I{\setpaddedsums;\Submodels[\noncuriousset] \given \gradpolyset^\mathcal{B}, \keypolyset^\mathcal{B},  \setgradientsums, \aggregatedgradient, \Randvecscol\!\!, \Submodels[\uecolluderset]}\nonumber \\
    &= \begin{aligned}[t] &\I{ \keypolyset^\mathcal{B},  \setgradientsums, \Randvecscol;\Submodels[\noncuriousset] \given \aggregatedgradient, \Submodels[\uecolluderset]} + \I{ \gradpolyset^\mathcal{B};\Submodels[\noncuriousset] \given \aggregatedgradient, \Submodels[\uecolluderset], \keypolyset^\mathcal{B},  \setgradientsums, \Randvecscol} \\
    &\!\! + \! \entropy{\setpaddedsums \given \gradpolyset^\mathcal{B}, \keypolyset^\mathcal{B},  \setgradientsums, \aggregatedgradient, \Randvecscol\!\!, \Submodels[\uecolluderset]} - \entropy{\setpaddedsums \given \Submodels[\noncuriousset], \gradpolyset^\mathcal{B}, \keypolyset^\mathcal{B},  \setgradientsums, \aggregatedgradient, \Randvecscol\!\!, \Submodels[\uecolluderset]}\end{aligned} \\
    &\overset{(b)}{=} \begin{aligned}[t] &\entropy{ \Submodels[\noncuriousset] \given \aggregatedgradient, \Submodels[\uecolluderset], \keypolyset^\mathcal{B},  \setgradientsums, \Randvecscol} - \entropy{ \Submodels[\noncuriousset] \given \gradpolyset^\mathcal{B}, \aggregatedgradient, \Submodels[\uecolluderset], \keypolyset^\mathcal{B},  \setgradientsums, \Randvecscol} \\
    &\!\!\! \overset{(c)}{+} \! (\vert \mainsets \vert - 1) \entropy{\randvec} - (\vert \mainsets \vert - 1) \entropy{\randvec} \end{aligned}\\
    &\overset{(d)}{=} 0,
\end{align*}
where (a) holds since $\aggregatedkeys$ by construction is a deterministic function of $\setgradientsums$ and (b) is because the keys are independent of the gradients and hence $\I{ \keypolyset^\mathcal{B},  \setgradientsums, \Randvecscol;\Submodels[\noncuriousset] \given \aggregatedgradient, \Submodels[\uecolluderset]} = 0$. We obtain (c) since the remaining entropy of $\setpaddedsums$ given $\gradpolyset^\mathcal{B}, \keypolyset^\mathcal{B},  \setgradientsums, \aggregatedgradient, \Randvecscol\!\!, \Submodels[\uecolluderset]$ is equal to the number of partial aggregations (minus one) times the full entropy of one key $\randvec$. This is because none of the partial sums in $\setpaddedsums$ can be decoded from any of the given information. By construction, there is always one unknown key remaining in the sum. However, the entire aggregation is known and hence one partial aggregation is deterministic given all others. The last equation (d) holds since additionally conditioning on $\gradpolyset^\mathcal{B}$ does not further reduce the entropy of $\entropy{ \Submodels[\noncuriousset] \given \aggregatedgradient, \Submodels[\uecolluderset], \keypolyset^\mathcal{B},  \setgradientsums, \Randvecscol}$ by the properties of secret sharing.
\end{proof}

\bibliographystyle{IEEEtran}
\bibliography{IEEEabrv,refs}

\end{document}